%% file: arxiv_cvpr_submission.tex
\documentclass[10pt,twocolumn,letterpaper]{article}

\usepackage[pagenumbers]{cvpr} %
\usepackage{indentfirst}

\input{preamble}

\definecolor{cvprblue}{rgb}{0.21,0.49,0.74}
\usepackage[pagebackref,breaklinks,colorlinks,citecolor=cvprblue]{hyperref}
\usepackage[dvipsnames]{xcolor}
\definecolor{pre_enroll_color}{HTML}{e0c487}
\definecolor{online_enroll_color}{HTML}{84B9D7}
\usepackage[dvipsnames]{xcolor}
\usepackage{tcolorbox}
\definecolor{pre_enroll_color}{HTML}{e0c487}
\definecolor{online_enroll_color}{HTML}{84B9D7}

\definecolor{darkblue}{rgb}{0,0,0.8}

\newcommand\datasheetSec[1]{\noindent \begin{tcolorbox}[width=0.9\linewidth, center] \centering {\bf #1}\end{tcolorbox}}
\newcommand\datasheetQ[1]{\noindent{\centering {\textcolor{darkblue}{#1}}} \\}

\newcommand{\answerYes}[1][]{{Yes.}}
\newcommand{\answerNo}[1][]{{No.}}
\newcommand{\answerNA}[1][]{{N/A}}

\def\paperID{2709} %
\def\confName{CVPR}
\def\confYear{2024}

\title{Instance Tracking in 3D Scenes from Egocentric Videos}

\author{Yunhan Zhao$^1$ \quad Haoyu Ma$^1$ \quad Shu Kong$^{2,3,4}$ \quad Charless Fowlkes$^1$ \\
$^1$UC Irvine  \quad $^2$Texas A\&M University \quad
$^3$Institute of Collaborative Innovation \quad $^4$University of Macau \\
{\tt\small \{yunhaz5, haoyum3, fowlkes\}@ics.uci.edu } \ \ {\tt\small skong@um.edu.mo}
\vspace{-2mm}
}

\begin{document}
\maketitle

\begin{abstract}

Egocentric sensors such as AR/VR devices capture human-object interactions and offer the potential to provide task-assistance by recalling 3D locations of objects of interest in the surrounding environment.
This capability requires instance tracking in real-world 3D scenes from egocentric videos (IT3DEgo).
We explore this problem by first introducing a new benchmark dataset, consisting of RGB and depth videos, per-frame camera pose, and instance-level annotations in both 2D camera and 3D world coordinates.
We present an evaluation protocol which evaluates tracking performance in 3D coordinates with two settings for enrolling instances to track:
(1) single-view online enrollment where an instance is specified on-the-fly based on the human wearer's interactions.
and (2) multi-view pre-enrollment where images of an instance to be tracked are stored in memory ahead of time.
To address IT3DEgo, we first re-purpose methods from relevant areas, e.g., single object tracking (SOT) --- running SOT methods to track instances in 2D frames and lifting them to 3D using camera pose and depth.
We also present a simple method that leverages pretrained segmentation and detection models to generate proposals from RGB frames and match proposals with enrolled instance images.
Our experiments show that our method (with no finetuning) significantly outperforms SOT-based approaches in the egocentric setting.
We conclude by arguing that the problem of egocentric instance tracking is made easier by leveraging camera pose and using a 3D allocentric (world) coordinate representation.
Dataset and open-source code: \href{https://github.com/IT3DEgo/IT3DEgo/}{ https://github.com/IT3DEgo/IT3DEgo}.
\vspace{-2mm}
\end{abstract}

\section{Introduction}
\begin{figure*}[t]
\centering
\hsize=\textwidth 
\includegraphics[width=0.99\textwidth]{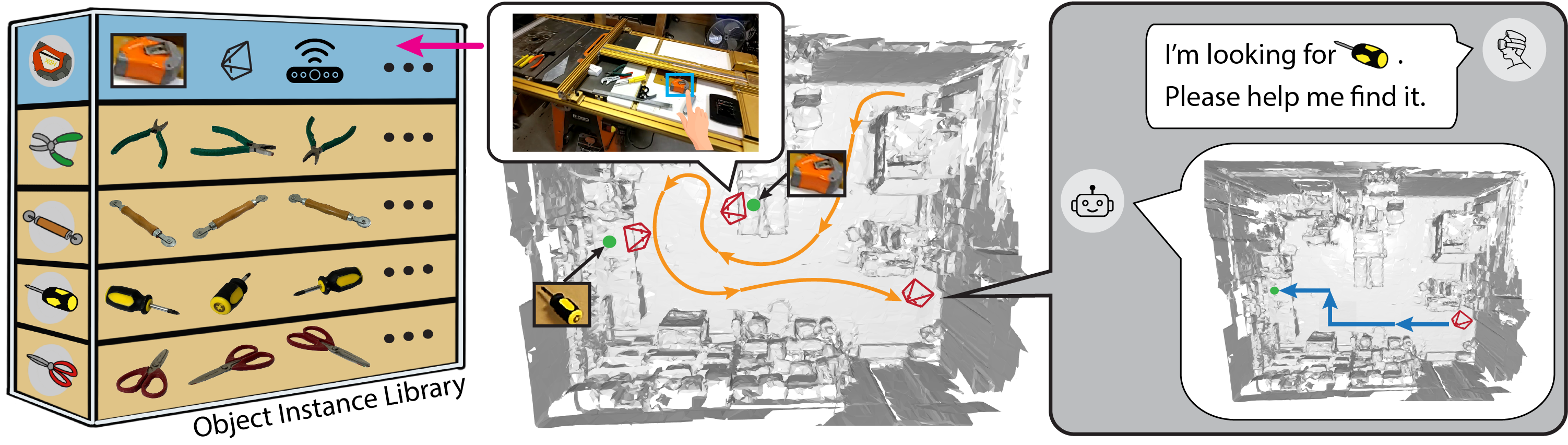}
\vspace{-2mm}
\caption{\small \textbf{Motivation for the proposed IT3DEgo benchmark task}. 
We envision the real-world application of an assistive agent that continuously tracks enrolled object instances in 3D and can provide navigation guidance to users to retrieve object instances at any time. 
Tracked objects are either  
{\setlength{\fboxsep}{0.pt}\colorbox{online_enroll_color}{\strut enrolled online}}
(first row in the library) where objects of interest are identified automatically based on user interactions or
{\setlength{\fboxsep}{0pt}\colorbox{pre_enroll_color}{\strut pre-enrolled}}
(bottom four rows in the library), where task-relevant objects are modeled from a collection of photos taken from different views. 
The former setup 
comes with additional in-context sensor information, such as camera pose and depth
while the latter 
features richer visual information.
}
\vspace{-3mm}
\label{fig:splash}
\end{figure*}

Egocentric video obtained from AR/VR devices provides a unique perspective that captures the interaction between the human wearer and the surrounding 3D environment. 
With the rapid development of AR/VR hardware, there is increasing interest in building assistive agents~\cite{nguyen2019help, szot2021habitat, wu2023tidybot, shi2023alexa}, that track the user's environment and provide contextual guidance on the location of objects of interest (illustrated in Figure~\ref{fig:splash}).
We argue that developing such an agent requires solving the largely unexplored problem of \emph{tracking object instances in 3D} from egocentric video.

{\bf Why this problem?}
First, tracking in egocentric video is a novel and underexplored problem, compared to the well-studied tracking from fixed, third-person viewpoints. More broadly, egocentric visual understanding tasks, such as human pose estimation and trajectory prediction~\cite{bi2020can, wang2022estimating, dunnhofer2023visual} are a growing area of interest.
Second, tracking in 3D scenes is essential in robotics, autonomous driving, and AR/VR applications.
Compared to the 2D counterpart, tracking objects in 3D is crucial for an agent to not only understand the surrounding 3D environment but also to determine precise locations for planning and navigation.
Combining the two perspectives above, there is a broader question of what information processing constraints govern how the human visual system integrates egocentric sensory data into a seemingly allocentric perception of the world around us.

{\bf Challenges and new opportunities.}
(1) Egocentric video often features motion blur, hand occlusions, and frequent object disappearances and reappearances which make the 2D tracking problem very challenging from pure visual signals~\cite{dunnhofer2023visual, tang2023egotracks}.
Tracking in 3D offers an opportunity to fuse additional sensor streams, such as depth and camera pose, to improve accuracy.
Unlike 2D tracking with a moving camera, 3D tracking in world coordinates allows the model to leverage the unique prior information -- {\it an object should remain still unless being interacted with the human operator}.
(2) 
For the downstream application of task guidance, we propose exploring novel approaches to identify or {\em enroll} object instances to be tracked.
One approach is automatically enrolling objects with which the user interacts or identifies via hand gestures such as pointing. Alternatively, object instances relevant to a particular task could be {\em pre-enrolled} based on a collection of images that specify the visual appearance of the object in advance.

{\bf Contribution 1: Dataset collection.} To our best knowledge, no existing dataset supports exploring the problem of IT3DEgo (c.f. Table~\ref{table:datasets_comparison}).
The recent Ego4D dataset~\cite{grauman2022ego4d} highlights some of these challenges. However, the Ego4D dataset only provides RGB frames\footnote{Ego4D does provide a sparse set of camera poses (less than 15\% of frames) estimated with COLMAP and predicted depth maps using monocular depth estimation.}
and sparse annotations (may miss potential object location changes), making it unsuitable to fully explore the problem.
We collect a new benchmark dataset with HoloLens2, including an RGB camera, a depth sensor, four grayscale cameras, per-frame camera pose and coarse scene geometry as a mesh. 
We describe the details of dataset statistics, capture procedures, and annotations in Section~\ref{sess:data_annot}.

{\bf Contribution 2: Benchmarking protocol.}
We propose a new IT3DEgo Benchmark for studying instance tracking in 3D scenes from egocentric videos with two settings for how objects are selected for tracking. 
(1) \emph{Tracking with single-view online enrollment (SVOE)} studies the scenario where object instances of interest are defined on-the-fly, i.e., objects are specified with a 2D bounding box in the frame where they {\it first} become fully visible to the user. 
(2) \emph{Tracking with multi-view pre-enrollment (MVPE)} assumes objects of interest are specified by multiple photos of the object of interest from different viewpoints before the tracking system starts. 
As detailed in Section~\ref{sess:protocol}, we evaluate performance with standard precision/recall metrics as well as geometric L2 and angular errors used in the Ego4D VQ3D evaluation~\cite{grauman2022ego4d}.

{\bf Contribution 3: Technical explorations.} 
Since our benchmark task is novel and underexplored in the literature, it is natural to re-purpose and evaluate existing approaches (e.g., SOT methods). 
We also explore an alternative {\it piecewise constant velocity} method that utilizes the Kalman filter~\cite{kalman1960new} with instance proposals from SAM~\cite{kirillov2023segment} and encoded by DINOv2~\cite{oquab2023dinov2}, resulting in drastic performance improvement over state-of-the-art SOT methods. Section~\ref{sess:methods} and~\ref{sess:experiments} provide details regarding baselines and benchmark results, respectively.
From the experimental results, we provide the following insights: Tracking object instances in egocentric videos is easier in 3D scenes leveraging camera poses and depth maps. Intuitively, an object not being interacted with has the same 3D position in a predefined world coordinate but the positions in 2D frames can change drastically due to the head motion. As a result, existing state-of-the-art 2D SOT approaches perform poorly on egocentric data. Future work should address the problem of re-identifying objects by leveraging the camera poses and accurately identifying and updating object motion changes.

{
\setlength{\tabcolsep}{2.7mm}
\begin{table*}[t]
\caption{\small \textbf{Comparisons of egocentric datasets that explore tracking-related problem.} 
Existing egocentric datasets only explore the tracking problem in 2D or predicting discrete 3D locations. Some mention the tracking problem in 3D but only consider limited sensor data (RGB) or synthetic environments.
Our benchmark dataset supports the study of instance tracking in 3D real-world scenarios (RWS in the table) from egocentric videos.}
\vspace{-3mm}
\small
\centering
\scalebox{0.9}{
\begin{tabular}{@{}lccccccccc@{}}
\toprule
\multirow{1}{*}{\bf Dataset} & {\bf Modality} & {\bf Device} & {\bf Avg. Length} & {\bf Annot. FPS} & {\bf RWS} & {\bf Camera Trajectory} & {\bf 3D Tracking} & {\bf Year} \\
\midrule
TREK-150~\cite{dunnhofer2021first} & RGB & GoPro & 10s & 60 & \cmark & Natural & \xmark  & 2021 \\
EK-VISOR~\cite{damen2018scaling} & RGB & GoPro & 12s & 0.9 & \cmark & Natural & \xmark & 2022 \\
Ego4D-VQ3D~\cite{grauman2022ego4d} & RGB & GoPro & - & - & \cmark & Natural & \xmark & 2022 \\
EMQA~\cite{datta2022episodic} & RGB-D+IMU & - & - & - & \xmark & Simulated & \xmark & 2022 \\
EgoPAT3D~\cite{li2022egocentric} &  RGB-D+IMU & Kinect & 4min & 30 & \xmark & Object-Centric & \xmark & 2022 \\
DigitalTwin~\cite{pan2023aria} & RGB-D+IMU & Aria & 2min & - & \xmark & Natural & \cmark & 2022 \\
EgoTrack~\cite{tang2023egotracks} & RGB & GoPro & 6min & 5 & \cmark & Natural & \xmark & 2022  \\
\midrule
{\bf Ours} & RGB-D+IMU & HoloLens & $>$5min & 6 & \cmark & Natural &\cmark & 2023 \\                
\bottomrule
\end{tabular}
}
\label{table:datasets_comparison}
\vspace{-3mm}
\end{table*}
}

\section{Related Work}

{\bf Egocentric video datasets} have been developed to study different problems over the last decade~\cite{lee2012discovering, pirsiavash2012detecting, fathi2012social, su2016detecting, damen2018scaling, grauman2022ego4d}. 
Traditionally, egocentric video understanding has focused on tasks such as activity recognition~\cite{ryoo2013first, possas2018egocentric, kazakos2019epic, plizzari2022e2}, human-object interactions~\cite{damen2014you, liu2020forecasting, liu2022hoi4d}, and inferring the camera wearer’s body pose~\cite{rogez2015first, jiang2017seeing, ng2020you2me, wang2022estimating}. 
Recently, more tasks have emerged due to the increasing interest in egocentric videos, such as action anticipation~\cite{rhinehart2017first, furnari2019would, fernando2021anticipating}, privacy protection~\cite{ryoo2017privacy, dimiccoli2018mitigating, thapar2021anonymizing}, and estimating social interactions~\cite{yonetani2016recognizing, li2019deep, northcutt2020egocom}. 
However, object tracking in egocentric videos is largely underexplored in the literature until the introduction of recent datasets\cite{dunnhofer2023visual, tang2023egotracks}.
These existing tracking datasets only support 2D tracking, which motivates us to collect and setup a new benchmark to evaluate real-world 3D instance tracking.

{\bf Tracking in 3D scenes} aims to identify objects of interest in 3D space from a sequence of frames. 
The prediction output format depends on the downstream tasks, including 3D bounding boxes~\cite{kim2021eagermot, weng20203d}, 3D object centers~\cite{zhou2019objects, yin2021center}, or 6DOF poses~\cite{garon2017deep, ahmadyan2021objectron}. 
State-of-the art 3D tracking models~\cite{zhou2022pttr, liu2023bevfusion, chen2023voxelnext} have focused on well-established third-person perspective benchmark datasets~\cite{geiger2013vision, caesar2020nuscenes, dai2017scannet}.
The recent large-scale Ego4D dataset starts to address the problem of querying the 3D positions of objects from a first-person perspective. 
However, the raw sensor data in Ego4D only includes RGB images and no other 3D information, such as depth and camera poses~\cite{zhao2020domain, zhao2021camera}. 
However, contemporary AR/VR headsets come with additional cameras, depth, and IMU sensors that allow for richer geometric reasoning~\cite{ungureanu2020hololens, pan2023aria}. Therefore, we believe it is realistic to leverage diverse sensor streams and explore the egocentric tracking problem in 3D.
Our benchmark dataset thus includes multiple raw sensors and derived data streams to support the study tracking in 3D scenes with modern hardware platforms.

{\bf Object instance detection and tracking} is a long-standing problem in computer vision and robotics~\cite{georgakis2016multiview, dwibedi2017cut, hodavn2019photorealistic, wang2020tracking, shen2023high}. Instead of predicting labels from a predefined set of object categories, instance-level predictions treat every object instance as a separate category. 
Instance-level tracking aims to locate given object instances in a sequence of frames, commonly using a tracking-by-detection paradigm. One common formulation is person re-identification~\cite{zheng2016person, ye2021deep}, which aims to track and associate individual people as they enter and leave multiple cameras' fields of view. Our setting is closely related but is dominated by the motion of the (egocentric) camera rather than the dynamics of object motion.

\begin{figure*}[t] 
\centering
\hsize=\textwidth 
\includegraphics[width=0.99\textwidth]{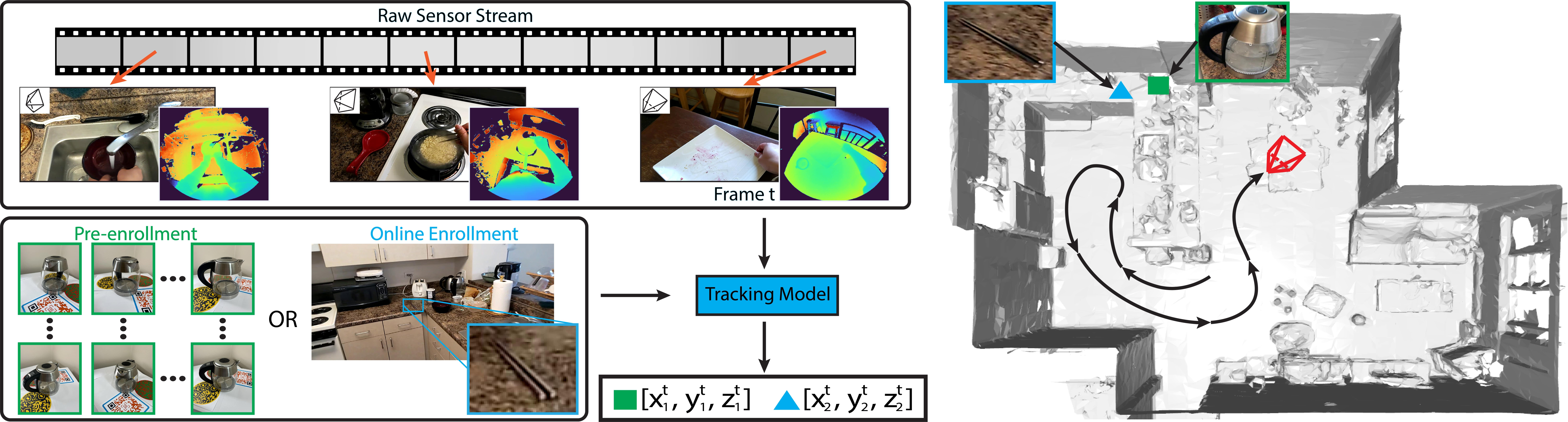}
\vspace{-2mm}
\caption{\small \textbf{Illustration of input and output of our benchmark task}.
Given a raw RGB-D video sequence with camera poses and object instances of interest, i.e., either by online enrollment (SVOE) or pre-enrollment (MVPE), the goal of our benchmark task is to output the object instance 3D centers in a predefined world coordinate at each timestamp. Please check Section~\ref{sess:protocol} for more details. 
}
\label{fig:input_output_pipeline}
\vspace{-3mm}
\end{figure*}

\section{IT3DEgo: Protocol and Dataset}

The problem of IT3DEgo is motivated by real-world assistive agents running on AR/VR devices.
Given an object instance specified by the end user, developed models are required to track it in the 3D environment, i.e., recording its 3D location over time (cf. Fig.~\ref{fig:input_output_pipeline}).
In this section, we introduce our benchmarking protocol and dataset.

\subsection{Benchmarking Protocol}
\label{sess:protocol}

Because object instances of interest are naturally diverse and may fall outside of the vocabulary of existing detectors, we set up a benchmarking protocol that focuses on evaluation without a separate training set. 
In other words, models should be pretrained on other data sources and cannot see objects in our dataset.
This aligns with the contemporary foundation models (e.g., CLIP~\cite{radford2021learning} and SAM~\cite{kirillov2023segment}) pretrained on open-world data.

{\bf Instances enrollment.} 
We consider two distinct setups to specify object instances of interest.
The first is {\it single-view online enrollment (SVOE)}, similar to single object tracking (SOT) where an object is specified on-the-fly by the end users.
For example, the user can specify an object of interest by interacting or pointing to it, after which the system should track it in the 3D world.
The second is {\it multi-view pre-enrollment (MVPE)}, which defines (or pre-enrolls) concerned objects with a set of object-centric images captured from multiple angles.
The two setups present different challenges.
SVOE provides a bounding box of the object (similar to specifying an object in SOT), 
but the visual quality is generally lower in resolution as the objects can be far from the camera.
MVPE provides 25 high-resolution (2124$\times$2832) object-centric images of the instances captured from different angles.
However, the object instance is captured under different lighting conditions than the tracking environment, and can be posed differently (e.g., keys can be deformed over time).

{\bf Evaluation protocols.}
Following the literature on object tracking and detection, we use the metrics below in our benchmarking protocol. 
\begin{itemize}[noitemsep,  topsep=1pt]
    \item {\bf Precision and recall} at different L2 distance thresholds. 
    Given $N$ specific thresholds $\tau_i$ with $i \in \{1, 2, .. N\}$, specifically 0.25, 0.5, 0.75, 1.0, and 1.5 meters, a ground-truth object location ${\bf o_{gt}} \in \mathbb{R}^3$ and a predicted location ${\bf o_{pred}} \in \mathbb{R}^3$, we count a true positive (TP$_{i}$) when $|| {\bf o_{gt}} - {\bf o_{pred}} ||_2 \leq \tau_i$. 
    At each timestamp, each ground-truth is matched to the prediction with the smallest L2 distance below the threshold. 
    Unmatched predictions and ground-truth at threshold $\tau_i$ are counted as false positives (FP$_{i}$) and false negatives (FN$_{i}$), respectively.
    TP$_{i}$, FP$_{i}$, and FN$_{i}$ are computed over all object instances in every frame. 
    The precision and recall at threshold $\tau_i$ is computed as $\sum$TP$_{i}$ / ($\sum$TP$_{i}$ + $\sum$FP$_{i}$) and $\sum$TP$_{i}$ / ($\sum$TP$_{i}$ + $\sum$FN$_{i}$), respectively~\cite{raghavan1989critical, yilmaz2006object}. 
    
    \item {\bf L2 and angular error.} Following VQ3D in Ego4D~\cite{grauman2022ego4d}, we also compute the L2 distance between the ground-truth and predictions in the world coordinates in meters.
    We also report the angular error in radians in the current camera coordinate system.
    Unlike threshold-aware 3D precision and recall, these metrics are computed only on frames where both ground-truth and prediction of the object instance location are available. 
\end{itemize}

To make 3D annotation tractable, we only evaluate predictions during time intervals when target objects are stationary (i.e., not being handled by the camera wearer).

\subsection{Dataset}
\label{sess:data_annot}
{\bf Raw video collection.}
The raw IT3DEgo data was recorded by three individuals in ten diverse indoor scenes, e.g., kitchen, garage, office, labs, etc.
The participants perform naturalistic tasks with different object instances in the scene, e.g., cooking, repairing, writing, etc. 
The raw data includes 50 recordings in total.
Each recording contains five or more object instances, each of which appears at three different 3D locations on average.
The average length of each recording is 10K frames or $>$5min.
We capture the raw data with HoloLens2 
(see Suppl.) 
which includes an RGB camera, a depth sensor, and four grayscale cameras with the resolution of 720$\times$1280, 480$\times$640, 512$\times$512, respectively.
Raw sensors operate at different frequencies, we sync all other sensors to the frequency of the RGB camera (30 fps).
We also provide a coarse resolution scene mesh of each environment reconstructed by the Hololense OS.
Additional details of the video sequences are described in the supplementary material.

{\bf Object instance collection.}
To support the SVOE setup, annotators identify the first RGB frame where a given object is fully visible and close enough to specify a 2D bounding box which is at least 500 pixels in area. For MVPE, we collected 25 high-resolution images of each object instance using iPhone 13 Pro. 
Each object was placed on a rotary table with QR codes. 
We took 12 photos of each object evenly from 360$^{\circ}$ while keeping the camera at about 30$^{\circ}$ elevation, 12 more at 60$^{\circ}$ elevation and 1 top-down view. We provide additional details and visualizations of object instances in the supplementary material.

{\bf Annotations.}
Our dataset includes three types of manual annotations: 
(1) {\it Object instance 3D centers} describe the 3D positions of each object instance center in a world coordinate frame. 
We annotate the 3D center by first averaging 3D points computed from camera poses and depth maps of different views of the object instance.
Annotators then examine and adjust computed 3D points by visualizing them together with the coarse mesh of the scene.
(2) {\it 2D bounding box annotations} are axis-aligned 2D bounding boxes of the instance every five frames starting from the beginning of the video. 
Specifically, we ask annotators to draw {\it amodal} bounding boxes of each object instance.
We do not annotate the object instances with heavy occlusions (i.e., when less than 25\% of the object is visible).
(3) {\it Object motion state annotations} are a per-frame annotation of whether the object is stationary or dynamic. For the data we collected, dynamic implies the camera wearer is interacting with the object.

\section{Methodology}
\label{sess:methods}

\subsection{Baseline: Re-purposed SOT Trackers}

To approach the problem of IT3DEgo,
we first explore a simple {\it unified pipeline} as the baseline approach based on single object tracking (SOT).
It allows instance-level 2D tracking by providing the visual appearance of object instances to track~\cite{danelljan2020probabilistic, bhat2019learning, yan2021learning}, which enables us to re-purpose them for our benchmark task.
In the unified pipeline, we first compute the 2D trajectories of each object instance with SOT.
The final 3D trajectories are computed by lifting the center of 2D bounding boxes with depth maps and camera poses.
Lastly, we adopt a simple memory mechanism that stores the previous locations of each object instance to handle the case where the instance moves out of sight, i.e., frames without valid predictions.

{\bf Lifting 2D trajectories to 3D.} 
With the 2D trajectory predicted from SOT, each valid 2D detection is then lifted into 3D space with the equation:
    ${\bf o}_{t}^{i} = {\bf T}_{t} z {\bf K}^{-1} {\bf c}^{i}_t$,
where ${\bf c}^{i}_t$ is the 2D coordinate of the center of the bounding box of instance $i$ at timestamp $t$, ${\bf o}_{t}^{i}$ is the 3D position of instance $i$ at timestamp $t$ in world coordinate. $z$ is the corresponding depth value of ${\bf c}^{i}_t$ on the depth map.
${\bf T}_{t}$ is the camera pose at timestamp $t$ that specifies the camera rotation and translation w.r.t to a predefined world coordinate. {\bf K} is the intrinsic matrix.
A frame may lack a valid 3D prediction because either there is no 2D location from SOT (e.g., the object is outside the field-of-view) or the depth map is missing the depth value at ${\bf c}^{i}_t$.

{\bf Completing 3D trajectories with memory.} 
Any given frame may lack a valid 3D prediction, either because there is no 2D location from SOT (e.g., the object is outside the field-of-view) or the depth map has missing depth values at ${\bf c}^{i}_t$.
To address this we implement a simple memory mechanism that stores only the most recent 3D location for each tracked instance (memory size=1). We update the memory whenever there is a new valid prediction. We note that this heuristic is a good match for the prior that object locations change only when they are being interacted with, in which case they should also be visible to the camera.

\subsection{Improved Baseline}
We also explore the approach that leverages the recent foundation model SAM~\cite{kirillov2023segment} and state-of-the-art feature encoder DINOv2~\cite{oquab2023dinov2} for IT3DEgo. 
Following a tracking-by-detection pipeline, we first compute the per-frame 2D detections of each object instance by comparing the cosine similarity of DINOv2 encoded features between candidate proposals from SAM and a visual feature template. Together with the depth and camera pose information, we convert the 2D detection of each object into a 3D point in a predefined world coordinate. A simple memory with size 1 is also adopted to handle the frames without valid predictions.

\begin{figure*}[t]
\centering
\hsize=\textwidth 
\includegraphics[width=0.99\textwidth]{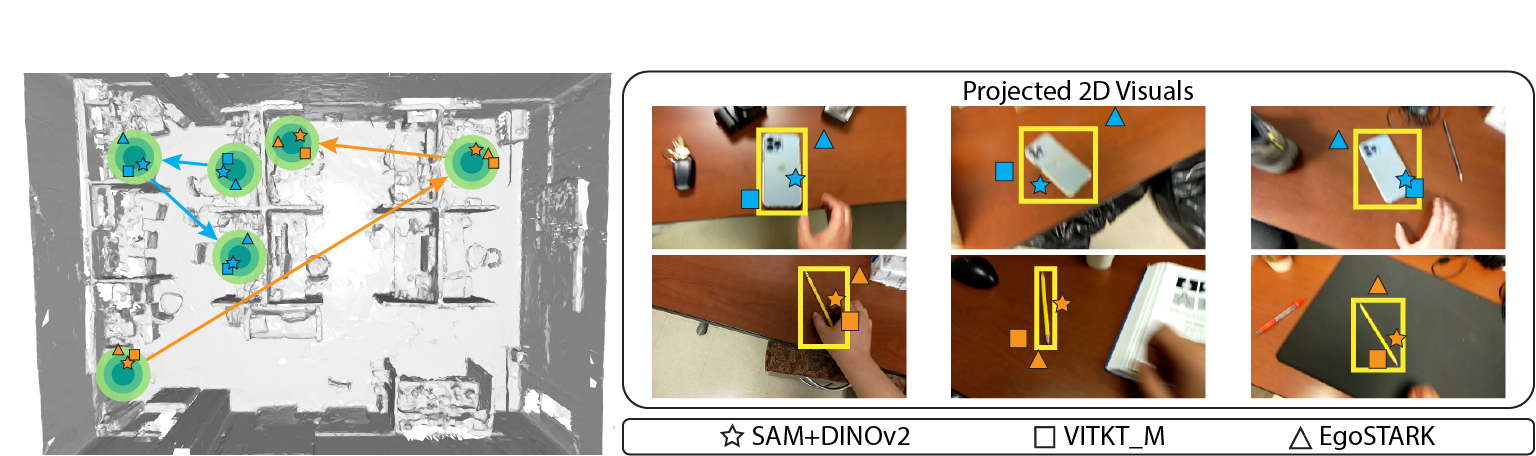}
\vspace{-1mm}
\caption{\small {\bf Qualitative visualizations of tracking with SVOE in both 3D space (left) and projected 2D view (right).} We visualize three top-performing trackers from different categories, i.e., EgoSTARK, VITKT\_M, and SAM+DINOv2. For projected 2D visualization, we compare the projected 3D points of each model w.r.t to the ground-truth annotated 2D bounding boxes. In the 3D view, we show 3 concentric circles at each ground-truth position representing 0.25, 0.5 and 0.75 meter thresholds. In both 2D and 3D visualizations, we find SAM+DINOv2 outperforms others as the predictions are closer to the center of object instances. 
}
\label{fig:qualitative_results}
\vspace{-3mm}
\end{figure*}

{\bf Exploring motion prior with Kalman filters.}
Currently, the naive update mechanism, i.e., always updating the memory for all incoming predictions, does not exploit the temporal information in video sequences. 
Inspired by the Kalman filter~\cite{weng2006video, patel2013moving, weng20203d} that is widely adopted in the tracking literature, we simply model the stationary position of each object instance as {\it piecewise constant velocity motion}, leveraging the prior information that an object without being interacted with has the same 3D coordinate.
Mathematically, the motion update with Kalman filter in each stationary position:
    $\hat{{\bf x}}_{t+1} = \hat{{\bf x}}_{t} + {\bf K}_{t}(\bf{z}_t - \bf{H}\hat{{\bf x}}_{t})$,
where $\hat{{\bf x}}_{t}$ is a 6DOF estimated state vector including position and velocity at time step $t$, ${\bf K}_{t}$ is the Kalman gain, $\bf{z}_t$ is a 3DOF the measurement vector, $\bf{H}$ is the observation matrix. Please refer to Kalman~\cite{kalman1960new} for more details.
Moving from one stationary position to the next one, we introduce an L2 distance heuristic to model the period where objects are being interacted.
Specifically, we compute the L2 distance between incoming 3D positions and the state predictions from the Kalman filter. 
If the L2 distance is above the threshold, we reset the Kalman filter with the current 3D predictions as the initialization.

\vspace{-2mm}
\section{Experiments}
\label{sess:experiments}

In this section, we first describe the implementation details of benchmark results. Then, we show the quantitative results of both setups and the visualizations of tracking results. Lastly, we demonstrate the importance of exploiting camera pose for tracking in 3D and perform ablation studies of the trackers. 
Note that we split our benchmark dataset into validation and test sets. All experiments are conducted on the validation set; the test set is used for future work.

{\bf Baseline SOT trackers.} 
We choose top-ranked trackers from well-established SOT literature and VOT challenges with open-source code for both tracking setups.
Specifically, we benchmark three short-term trackers ToMP~\cite{mayer2022transforming}, MixFormer~\cite{cui2022mixformer}, and ARTrack~\cite{wei2023autoregressive}; and three top-performing trackers from VOT long-term tracking challenges 2021~\cite{Kristan2021a} and 2022~\cite{Kristan2022a}, mixLT, mlpLT and VITKT\_M. We also evaluate trackers that utilize additional depth information as part of the input, including SAMF and MixForRGBD from VOT RGB-D tracking challenge 2022~\cite{Kristan2022a}, and ViPT~\cite{zhu2023visual}. Lastly, we benchmark the recent egocentric specific finetuned trackers, EgoSTARK~\cite{tang2023egotracks}.
Note that SOT trackers require initial bounding boxes to track, which are not available in MVPE. When re-purposing to MVPE setup, we explore two different initializations: 
(1) {\it detection-based initialization:} use multi-view pre-enrollment images to search for the initial bounding boxes where object instances first appear in the video and initialize SOT trackers with the predicted 2D boxes.
(2) {\it template-based initialization:} directly adopt multi-view pre-enrollment images as visual templates in the tracker and set the initial tracking search region to the entire frame. 

{\bf Implementation details.} 
The cosine similarity threshold in the SAM+DINOv2 approach is 0.6, i.e., the object is considered not visible if the cosine similarity is smaller than the threshold. 
For a fair comparison, we add additional 2D prediction filtering when re-purposing SOT trackers. We discard 2D predictions from SOT trackers whose prediction scores are lower than 75\% of the maximum prediction score.
When tracking with MVPE, we first preprocess the captured multi-view images by segmenting and cropping the foreground object using~\cite{li2023matting}.
Many transformer-based SOT trackers only encode a limited number of templates, therefore, we choose 5 images from 0$^\circ$, 90$^\circ$, 180$^\circ$, 270$^\circ$ and top-down for all models in MVPE experiments. 
We include an ablation study exploring the relationship between tracking results and the number of views used in the supplementary material.
To keep the comparison fair, all detection-based trackers in MVPE use SAM+DINOv2 with the same cosine thresholds to locate the initial bounding boxes.
In terms of benchmarking RGB-D trackers in MVPE, we utilize the estimated sparse depth maps using COLMAP~\cite{schoenberger2016sfm}.
The L2 distance threshold of resetting Kalman filters is 0.15m.
All experiments are implemented with PyTorch and run on Nvidia 2080Ti GPUs.

{
\setlength{\tabcolsep}{3.1mm}
\begin{table*}[t]
\footnotesize
\centering
\caption{\small \textbf{Benchmark results of tracking with SVOE. 
} From the results, we draw three salient conclusions: (1) The ability of re-identifying object instances after they disappear is important, as long-term and egocentric specific trackers outperform short-term trackers, i.e., RGB-ST and RGB-D. 
(2) Currently, encoding depth maps as auxiliary information cannot improve performance since depth maps are sparse and not always perfectly aligned with RGB frames due to distortions.
(3) The Kalman filter smoothing yields marginal improvements over the simple memory heuristic.
The method with KF subscript indicates it applies the Kalman filter.
}
\vspace{-3mm}
\scalebox{0.95}{
\begin{tabular}{lccccccccccccc}
\toprule 
\multirow{2}{*}{Model} & \multirow{2}{*}{Modality} & \multicolumn{5}{c}{Precision(\%)$\uparrow$} & \multicolumn{5}{c}{Recall(\%)$\uparrow$} & \multirow{2}{*}{L2$\downarrow$} & \multirow{2}{*}{Angle$\downarrow$} \\
\cmidrule(l){3-7} \cmidrule(l){8-12}
& & 0.25 & 0.5 & 0.75 & 1.0 & 1.5 & 0.25 & 0.5 & 0.75 & 1.0 & 1.5 & (m) & (rad) \\
\midrule
ToMP & RGB-ST & 5.6 & 10.1 & 17.2 & 25.3 & 39.0 & 6.1 & 11.0 & 18.8 & 27.7 & 42.6 & 2.11 & 1.32 \\
MixFormer & RGB-ST & 8.3 & 12.2 & 18.7 & 27.0 & 43.0 & 9.0 & 13.4 & 20.4 & 29.5 & 47.0 & 1.97 & 1.15 \\
ARTrack & RGB-ST & 9.1 & 13.9 & 21.5 & 30.3 & 45.1 & 10.1 & 15.3 & 23.7 & 32.4 & 47.2 & 1.92 & 1.10 \\
\midrule
SAMF & RGB-D & 7.0 & 11.5 & 15.7 & 24.0 & 40.8 & 7.7 & 12.5 & 17.2 & 26.3 & 44.7 & 1.90 & 1.00 \\
MixForRGBD & RGB-D & 7.5 & 12.1 & 16.8 & 25.3 & 41.0 & 8.3 & 13.4 & 20.1 & 28.5 & 45.0 & 2.11 & 1.32 \\
ViPT & RGB-D & 8.9 & 13.6 & 20.6 & 28.1 & 41.4 & 9.7 & 14.9 & 22.5 & 30.7 & 45.3 & 2.02 & 1.21 \\
\midrule
mixLT & RGB-LT & 14.4 & 17.5 & 23.9 & 31.8 & 47.2 & 15.8 & 19.2 & 26.1 & 34.8 & 51.6 & 1.85 & 1.02 \\
mlpLT & RGB-LT & 16.0 & 20.0 & 25.5 & 35.2 & 48.2 & 16.7 & 20.8 & 26.5 & 36.7 & 50.1 & 1.77 & 0.97 \\
VITKT\_M & RGB-LT & 21.5 & 24.2 & 29.7 & 37.5 & 50.6 & 23.0 & 25.9 & 31.8 & 40.2 & 54.2 & 1.55 & 0.83 \\
\midrule
EgoSTARK & RGB-Ego & 17.5 & 21.2 & 26.8 & 36.3 & 49.1 & 17.6 & 22.0 & 27.4 & 38.0 & 51.2 & 1.70 & 0.91 \\
\midrule
SAM + DINOv2 & RGB & 23.3 & 26.4 & 33.1 & 43.3 & 59.4 & 24.9 & 28.1 & 35.3 & 46.3 & 63.4 & 1.35 & 0.81 \\
SAM + DINOv2$_{\text{KF}}$ & RGB & {\bf 23.7} & {\bf 27.1} & {\bf 33.9} & {\bf 44.5} & {\bf 61.2} & {\bf 25.5} & {\bf 29.0} & {\bf 36.8} & {\bf 48.0} & {\bf 64.9} & {\bf 1.32} & {\bf 0.79} \\
\bottomrule
\end{tabular}
}
\vspace{-2mm}
\label{tab:benchmark_results_SVOE}
\end{table*}
}

\subsection{Benchmark Results}

{\bf Tracking with SVOE.}
From the results shown in Table~\ref{tab:benchmark_results_SVOE}, we have the following salient insights:
(1) {\it Re-identifying object instances is important.} Trackers designed with strong re-identify ability, i.e., long-term and egocentric specific types, outperform short-term trackers.
Similar findings are shown in recent 2D egocentric tracking work~\cite{dunnhofer2023visual, tang2023egotracks}. %
Surprisingly, SAM+DINOv2, the non-learned approach which does not exploit temporal information beyond the memory heuristic, performs the best among all baselines. We believe the exhaustive proposals on every frame and high quality features provide the model strong, generic re-identification ability.  
(2) {\it Depth information is not fully leveraged.} Current RGB-D trackers show similar or slightly worse performance compared to RGB-ST trackers (c.f. MixFormer and MixforRGBD). The main reason is that RGB-D trackers only encode depth maps as auxiliary visual features, which cannot fully exploit the geometric information from depth maps. Additionally, the depth maps are sparse and not always perfectly aligned with RGB images due to camera distortions. 
(3) {\it Simple Kalman filter brings marginal benefits.}
The Kalman filter does not improve over the simple ``most recent'' memory heuristic for stationary objects. The naive filter is also not sufficient for modeling the switching between stationary and dynamic motions needed to capture user-object interactions.

{\bf Tracking with MVPE.} 
We benchmark top-performing trackers in each category in Table~\ref{tab:benchmark_results_SVOE} for MVPE setup. 
From the results shown in Table~\ref{tab:benchmark_results_MVPE}, we find:
(1) {\it SOT trackers cannot fully exploit pre-enrollment information.} 
SOT methods rely on the initial position defined by 2D boxes on the frame to perform well.
Comparing detection-based initializations and SVOE results, e.g., ARTrack$^{\text{D}}$ and ARTrack in Table~\ref{tab:benchmark_results_SVOE}, the model performance drops since the initial boxes are not as accurate as ground-truth initialization.
VITKT\_M adopts many complicated modules that all rely on the initial bounding boxes and degrades more significantly, compared to other types of trackers. 
(2) {\it Encoding rich visual information generally helps.}
From the results of template-based initializations, VITKT\_M for the same reason mentioned before, we find trackers benefit from the high-resolution multi-view images. 
SAM+DINOv2 shows a significant performance boost because it is more robust to inaccurate initialization without relying on temporal information.

{\bf Qualitative results.}
Figure~\ref{fig:qualitative_results} shows predictions of top-performing trackers from three different categories, i.e., best tracker in long-term and egocentric specific, and SAM+DINOv2.
Clearly, SAM+DINOv2 predictions are closer to the object center in both 3D and projected 2D space.

\subsection{Further Analysis and Ablation Study}
\label{sess:analyses_and_ablation}
We further compare tracking object instances in both 2D and 3D settings, demonstrating tracking object instances is much easier in 3D space. 
We also include an ablation study regarding the cosine similarity thresholds.
All studies shown in this section are using SAM+DINOv2 unless otherwise specified.
More quantitative and qualitative results are shown in the supplementary materials.

{
\setlength{\tabcolsep}{3.1mm}
\begin{table*}[t]
\footnotesize
\centering
\caption{\small \textbf{Benchmark results of tracking with MVPE.
}
We evaluate top-performing trackers in each category in Table~\ref{tab:benchmark_results_SVOE} for MVPE setup.
From the results, we have the following summaries: 
(1) {\it SOT trackers cannot fully exploit pre-enrollment information.} 
Detection-initialized versions perform less well compared to SVOE due to the inaccurate estimated initial bounding boxes. VITKT\_M, which uses many modules that rely heavily on the initialization, degrades more significantly.
(2) {\it Encoding rich visual information generally helps.}
SAM+DINOv2 shows an even larger performance boost because it is more robust to the inaccurate initialization.
The D and T superscripts indicate the detection- and template-based initializations, respectively.
}
\vspace{-3mm}
\scalebox{0.95}{
\begin{tabular}{lccccccccccccc}
\toprule 
\multirow{2}{*}{Model} & \multirow{2}{*}{Modality} & \multicolumn{5}{c}{Precision(\%)$\uparrow$} & \multicolumn{5}{c}{Recall(\%)$\uparrow$} & \multirow{2}{*}{L2$\downarrow$} & \multirow{2}{*}{Angle$\downarrow$} \\
\cmidrule(l){3-7} \cmidrule(l){8-12}
& & 0.25 & 0.5 & 0.75 & 1.0 & 1.5 & 0.25 & 0.5 & 0.75 & 1.0 & 1.5 & (m) & (rad) \\
\midrule
ARTrack$^{\text{D}}$ & RGB-ST & 6.8 & 12.1 & 18.5 & 25.8 & 41.0 & 7.1 & 12.3 & 18.8 & 26.8 & 42.1 & 1.98 & 1.16 \\
ARTrack$^{\text{T}}$ & RGB-ST & 11.2 & 18.1 & 25.2 & 28.7 & 38.5 & 12.7 & 20.9 & 28.5 & 33.0 & 45.1 & 1.91 & 1.07 \\
\midrule
ViPT$^{\text{D}}$ & RGB-D &  6.3 & 11.7 & 17.9 & 24.9 & 40.2 & 6.9 & 11.8 & 17.9 & 25.9 & 41.0 & 2.01 & 1.21 \\
ViPT$^{\text{T}}$ & RGB-D &  10.5 & 17.4 & 24.0 & 27.1 & 36.3 & 11.9 & 20.1 & 27.1 & 31.3 & 44.0 & 1.93 & 1.10 \\
\midrule
VITKT\_M$^{\text{D}}$ & RGB-LT & 13.8 & 18.0 & 25.0 & 33.0 & 46.6 & 14.3 & 18.6 & 25.8 & 34.1 & 48.2 & 1.77 & 0.98 \\
VITKT\_M$^{\text{T}}$ & RGB-LT &  9.2 & 12.8 & 20.7 & 28.5 & 44.0 & 9.7 & 14.2 & 22.4 & 31.3 & 46.5 & 1.95 & 1.08 \\
\midrule
EgoSTARK$^{\text{D}}$ & RGB-Ego & 13.2 & 17.0 & 23.1 & 30.5 & 47.0 & 14.7 & 18.5 & 25.9 & 34.4 & 49.7 & 1.82 & 1.01 \\
EgoSTARK$^{\text{T}}$ & RGB-Ego & 18.9 & 23.1 & 28.3 & 37.1 & 49.6 & 19.1 & 23.1 & 29.3 & 39.1 & 52.9 & 1.67 & 0.88 \\
\midrule
SAM + DINOv2 & RGB & 56.0 & 59.0 & 61.8 & 67.5 & 74.3 & 50.0 & 52.7 & 55.2 & 60.3 & 66.4 & 0.67 & 0.40 \\
SAM + DINOv2$_{\text{ KF}}$ & RGB & {\bf 56.2} & {\bf 59.4} &{\bf 62.2} & {\bf 68.1} & {\bf 74.8} & {\bf 50.3} & {\bf 53.1} & {\bf 55.7} & {\bf 61.1} & {\bf 67.1} & {\bf 0.65} & {\bf 0.39} \\
\bottomrule
\end{tabular}
}
\vspace{-3mm}
\label{tab:benchmark_results_MVPE}
\end{table*}
}

\begin{figure*}[t]
\centering
\hsize=\textwidth 
\includegraphics[width=0.99\textwidth]{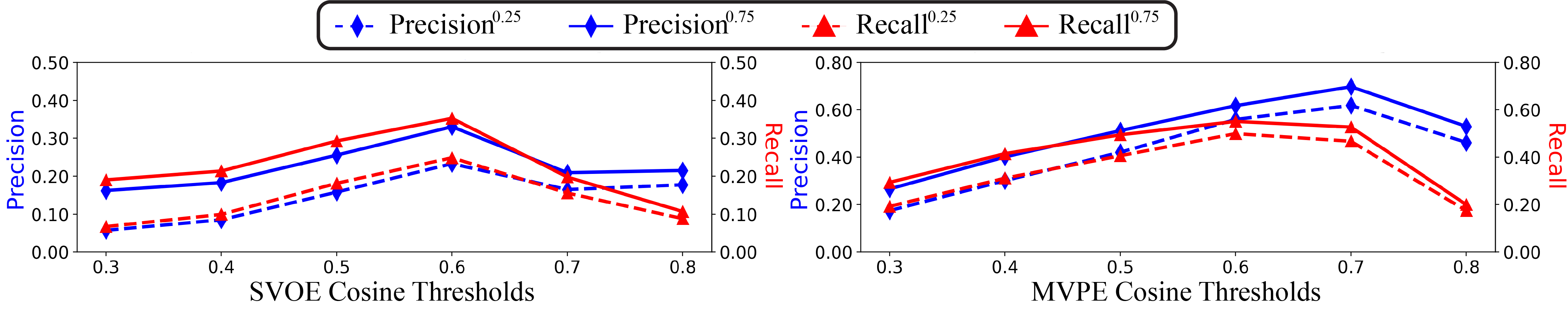}
\vspace{-3mm}
\caption{\small {\bf Performance comparisons of SAM+DINOv2 with different cosine thresholds.} By increasing the threshold, we find the model performance first improves and then gradually decreases. 
Intuitively, increasing the threshold will initially filter noisy predictions but when the threshold is too large the model will miss correct object 3D location updates.}
\label{fig:cos_thresh}
\vspace{-4mm}
\end{figure*}

{
\setlength{\tabcolsep}{2.8mm}
\begin{table}[t]
\centering
\footnotesize
\caption{\small \textbf{Quantitative comparisons of 2D tracking results w/ and w/o 3D guidance.}
With 3D guidance means the 2D results are computed by finding the bounding box proposal with the smallest L2 distance from projected 3D trajectories. Without 3D guidance means proposals are selected purely based on the visual feature cosine similarity. Please refer to Section~\ref{sess:analyses_and_ablation} for more details. From the results, we find the tracking results are significantly improved with the 3D guidance, indicating that tracking in 3D in egocentric videos is much easier than in 2D by leveraging camera pose and depth sensors. 
}
\vspace{-2mm}
\scalebox{0.95}{
\begin{tabular}{ccccc}
\toprule
 & 3D Guid. & AUC(\%)$\uparrow$ & N. Prec.(\%)$\uparrow$ & Prec.(\%)$\uparrow$ \\
\midrule
\multirow{2}{*}{SVOE} & \xmark & 20.7 & 14.9 & 8.9 \\
 & \checkmark & \bf{27.6} & \bf{21.7} & \bf{11.5} \\
\midrule
\multirow{2}{*}{MVPE} & \xmark & 14.1 & 7.4 & 3.0 \\
 & \checkmark & \bf{39.1} & \bf{35.2} & \bf{18.7} \\
\bottomrule
\end{tabular}}
\label{tab:2d_vs_3d_tracking}
\vspace{-4mm}
\end{table}
}

{\bf Tracking in 2D with 3D guidance.}
We experimentally demonstrate the importance of leveraging 3D information in egocentric instance tracking by comparing 2D tracking results w/ and w/o 3D guidance.
With 3D guidance means the 2D tracking results are computed with {\it predicted} 3D trajectories as the guidance.
For each object instance, the per-frame 2D detection results are computed by selecting the (above threshold) proposal with the smallest L2 distance between projected 3D points and the center of bounding boxes proposals. 
Without 3D guidance means the 2D tracking results are produced by selecting the proposal with the highest cosine feature similarity.
To keep a fair comparison, the cosine similarity threshold is the same when computing the 3D and 2D trajectories.
We evaluate the 2D tracking performance using widely adopted precision, normalized precision metrics and AUC in SOT literature~\cite{wu2013online}.
As shown in Table~\ref{tab:2d_vs_3d_tracking}, the model with 3D guidance performs significantly better in both SVOE and MVPE, demonstrating that leveraging the 3D information, such as camera pose and depth map, makes the tracking problem much easier.

{\bf Performance w.r.t cosine similarity thresholds.} 
Feature cosine similarity threshold is adopted to determine whether the object instance is present in the current frame, which is crucial for the memory updating mechanism.  
To characterize the relationship between tracking performance and cosine similarity thresholds, we run the experiment with different cosine similarity thresholds but keep everything else the same.
As shown in Figure~\ref{fig:cos_thresh}, both models show improved performance at first and then a gradual decrease. Higher cosine thresholds result in fewer predictions so the model must increasingly rely on previous confident predictions stored in the memory. 
Models with large cosine similarity thresholds have a higher chance of missing valid location updates, which leads to a drop in both precision and recall.

\section{Discussion}
\label{sect:diss}

{\bf Limitations and future work.} 
We point out that the current benchmark dataset has limited geographic and demographic diversity and captures only a small range of objects and activities. As such it is not appropriate for training large models and only serves as a diagnostic test to identify some limitations of existing approaches. Our hope is that it serves as a starting point for the research community to explore and eventually grow into a more comprehensive challenge.
Currently, the studied baseline approaches follow the same paradigm, i.e., lifting predicted 2D trajectories into 3D space. We found empirically that the simplest memory mechanism performed best but it seems very likely there are more nuanced state-update models which can integrate multiple observations effectively. 

Finally, we highlight two opportunities for future work.
First, advanced models to detect object 3D motion changes. 
Our experiments demonstrate that tracking in 3D world coordinates effectively narrows the problem to that of accurately predicting the object motion status, i.e., finding all stationary periods for each object instance. However, accurately predicting object state changes is still a non-trivial problem to solve.
Second, better utilization of object instance information. Currently, the object instances enrollments, i.e., SVOE and MVPE are naively encoded as visual features. Future work should explore the approaches of fusing the additional scene 3D information with object instances for better tracking performance.

{\bf Broader impact.} We believe the broader impact of our work is two-fold. 
First, we hope our benchmark brings more attention to the problem of tracking object instances in 3D from the egocentric perspective and contributes towards building future task-aware assistive agents.
Second, our multi-modal benchmark dataset is beneficial to the study of other 3D scene understanding related problems from the egocentric perspective, such as SLAM, camera localization, 3D reconstruction, and depth estimation.

{\bf Potential negative impacts.} Tracking in 3D from egocentric videos requires the geometric data of surrounding environments and the sensor streams that continuously capture their workplace or daily lives. There are obvious privacy concerns when deploying such hardware and algorithms. Similar to other apps running on personal devices, the simple solution is to keep all user data locally or (in the context of research) develop techniques for anonymizing video~\cite{thapar2021anonymizing}.

\section{Conclusion}
We introduce a new {\em IT3DEgo} benchmark that allows us to study the problem of tracking object instances in 3D from egocentric videos. The object instances to be tracked are either determined in advance or enrolled online during user interactions with the environment.
To support the study, we collect and annotate a new dataset that features RGB-D videos and per-frame camera poses, along with instance-level annotations in both 2D camera and 3D world coordinate frames.
We re-purpose and evaluate state-of-the-art single object trackers and develop a strong baseline using large pretrained recognition models and Kalman filtering.
We hope our benchmark brings more attention to this challenge and contributes to the development of perceptually-aware assistive agents.

{\bf Acknowledgements.} This work was partially supported by the DARPA Perceptually enabled Task Guidance (PTG) Program (HR00112220005). Shu Kong was partially supported by the University of Macau (SRG2023-00044-FST).
We thank Juan Carlos Dibene for the help of using hl2ss~\cite{hl2ss}; An Dang for the help with the initial development of the data capture; Yi-Ling Lin for the help of partial data annotations.

{
    \small
    \bibliographystyle{ieeenat_fullname}
    \bibliography{main}
}

\clearpage
\appendix
\addcontentsline{toc}{section}{Appendices}

\begin{center}
{\bf \Large Appendices}
\end{center}

\begin{center}
{\bf \large Outline}
\end{center}
This document supplements the main paper with additional details of the benchmark dataset, more experimental results, dataset documents, and visualizations. 
Below is the outline of this document.

\begin{itemize} [noitemsep, topsep=-1pt, leftmargin=*]    

\item 
    {\bf Section~\ref{sec:details of the dataset}}. Additional details of the benchmark dataset, including the capture procedure of raw videos and object instances, and the annotation steps.

\item 
    {\bf Section~\ref{sec:ablation}}. We conduct extensive studies and analyses on the improved baseline approach, SAM+DINOv2.

\item 
    {\bf Section~\ref{sec:datasheets for datasets}}. Dataset documentation and intended uses.
    
\item 
    {\bf Section~\ref{sec:vis}}. Visualizations of 2D frames from raw video sequences and 3D meshes of the capture environments. 
    
\end{itemize}

\section{Additional Dataset Details}
\label{sec:details of the dataset}
We present additional details of the datasets, such as collection details and annotations, to help others better understand and utilize the benchmark dataset.
Note that the data collection protocol was registered with the appropriate institutional review board (IRB).

\begin{figure}[t]
\includegraphics[width=1\linewidth]{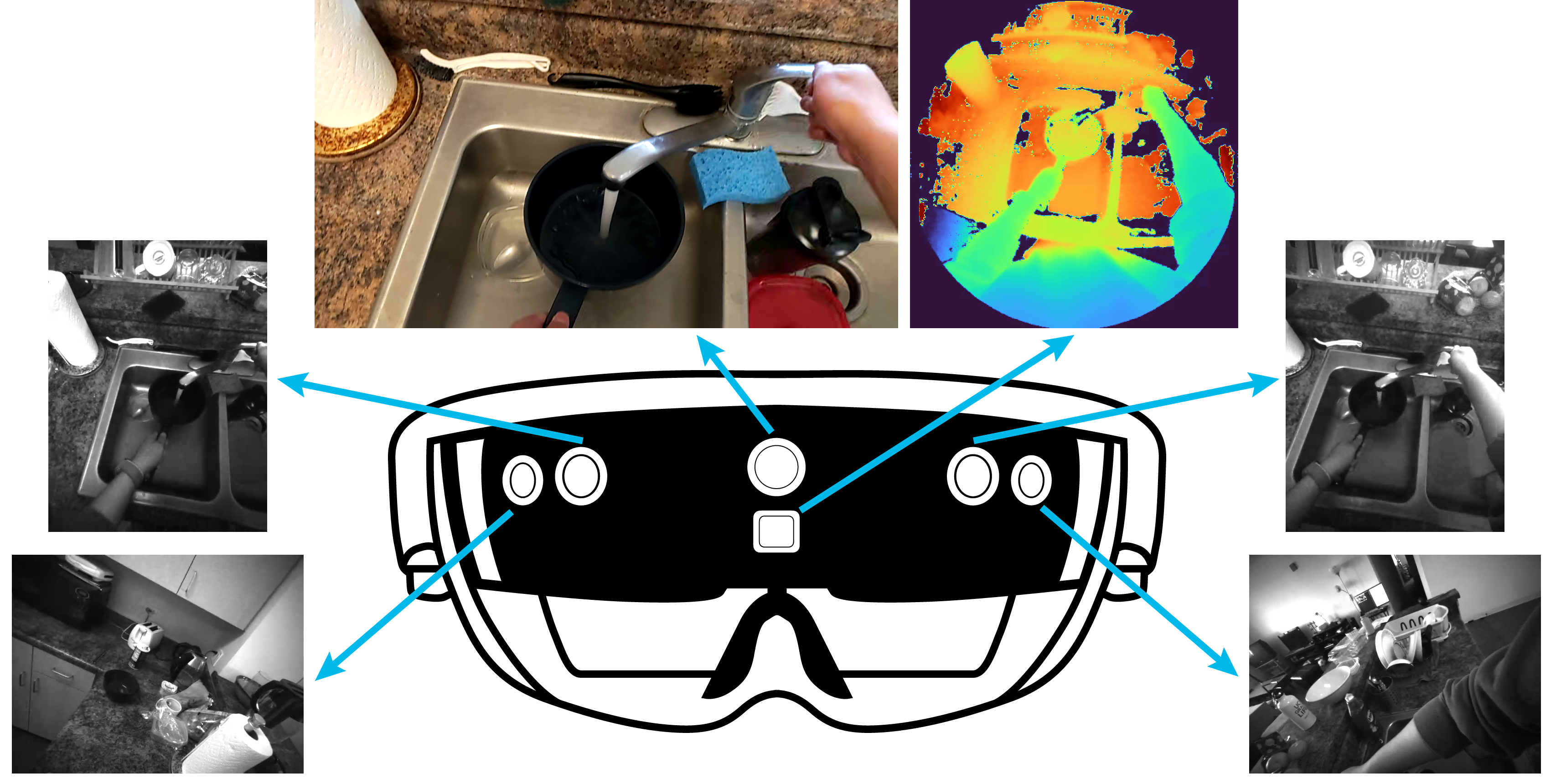}
\vspace{-4mm}
\caption{\small \textbf{Illustration of our benchmark dataset.} 
It is collected with HoloLens2 which captures RGB, depth, and four grayscale side views at 30 fps. Additionally, the device also captures per-frame camera poses allowing coarse reconstruction of the surroundings.}
\vspace{-3mm}
\label{fig:data_visual}
\end{figure}

{\bf Raw video collections.}
We capture the raw data using HoloLens2 that includes 1 RGB camera, 4 grayscale cameras, and 1 depth sensor operating in 2 different modes, shown in Figure~\ref{fig:data_visual}. 
Considering the downstream application scenarios of our benchmark task, we choose to capture our benchmark dataset in 10 different indoor scenes.
To capture the real-time geometry information, we capture all videos with high fps AHAT depth mode in HoloLens2~\cite{ungureanu2020hololens}.
Note that AHAT depth maps come with phase wrapping~\cite{hansard2012time} at 1 meter but they can be unwrapped using rendered depth from mesh or exploring existing unwrapping algorithms~\cite{droeschel2010probabilistic, droeschel2010multi}.
Before capturing in a new environment, we have a warm-up phase to make the device familiar with the surrounding environment in order to output accurate camera poses when capturing the video. 
In the warm-up phase, we walk around in the environment with the HoloLens2 turned on and make sure the device has seen all visible surfaces.
In practice, we spend around 20 minutes for the warm-up phase when we move to a new environment and around 5 minutes every time before we capture the new video. 

\begin{figure*}[t]
\centering
\hsize=\textwidth 
\includegraphics[width=0.99\textwidth]{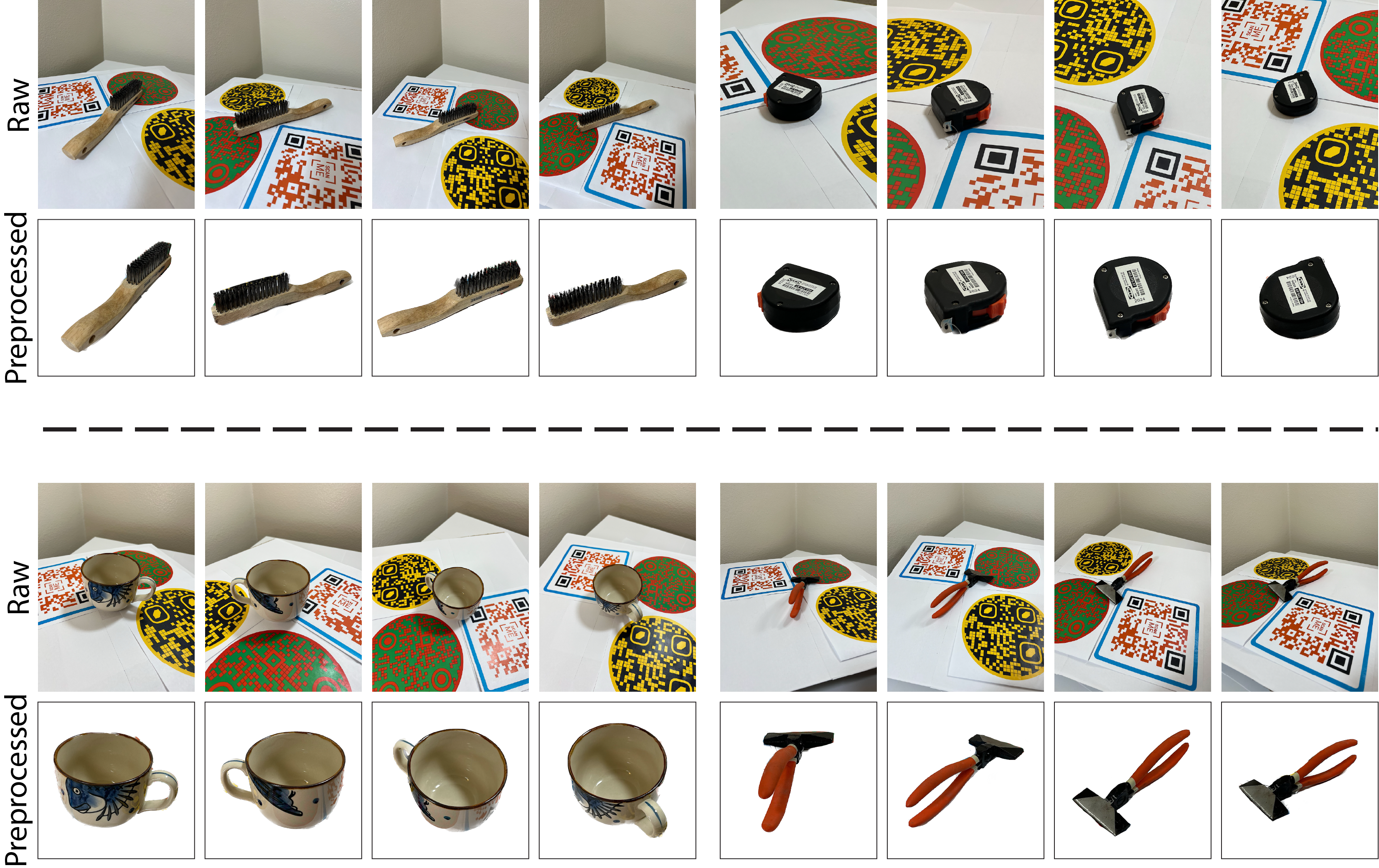}
\vspace{-2mm}
\caption{\small \textbf{Visualization of raw and preprocessed multi-view images}. 
Raw images represent the images directly output from the capture device, i.e., iPhone 13 Pro. We process raw images with segmentation and cropping before feeding them into the models. For more implementation details, please check Section 5 in the main paper.
}
\vspace{-2mm}
\label{fig:multiview_visual}
\end{figure*}

\begin{figure}[t]
\includegraphics[width=1\linewidth]{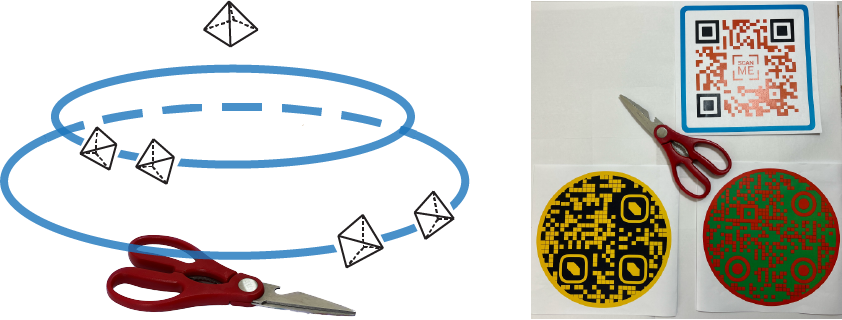}
\vspace{-4mm}
\caption{\small \textbf{Illustration of our multi-view capture setup.} 
The left panel shows our camera positions when taking 25 images to support the pre-enrollment study. Specifically, we take 12 object-centric photos evenly from 360$^\circ$ while keeping the camera 30$^\circ$ elevation. Another 12 images are taken in a similar fashion while keeping the camera 60$^\circ$ elevation. Lastly, we take one top-down view. An example of the top-down view with the QR code is shown on the right. 
}
\vspace{-6mm}
\label{fig:multiview_capture}
\end{figure}

{\bf Object instance collections.}
The entire videos come with 220 unique object instances, which cover a wide range of object instances for naturalistic daily tasks, such as cooking, writing, and repairing.
For each instance, we take 25 high-resolution images on a rotary table with the QR code (c.f. Figure~\ref{fig:multiview_visual} for visual examples).
Specifically, the photos are taken by hand-held iPhone 13 Pro approximately 45 cm away from the object center.
As illustrated in Figure~\ref{fig:multiview_capture}, we took 12 photos of each object evenly from 360$^{\circ}$ while keeping the camera at about 30$^{\circ}$ elevation, 12 more at 60$^{\circ}$ elevation and 1 top-down view.
We zoom in 2.5 times for objects whose diameter is lower than 20 cm to ensure the object instance is large enough in the image and use the normal scale (no zoom) for the rest of the case.

{\bf Annotations.}
There are three types of manual annotations along with our benchmark dataset. 
First, the 3D center of each object instance. 
The annotator is first asked to draw boxes on depth maps from $\geq$5 diverse views if possible. 
Each 2D bounding box is lifted to the 3D space with camera poses. The 3D centers of each object instance in a stationary
period are averaged to get the initial estimation.
The annotators then examine the adjust the annotated 3D points based on the RGB frames from the video sequence and captured mesh.
Second, the 2D axis-align bounding boxes of each object instance every five frames starting from the beginning of the video. 
Specifically, we ask the annotators to go through the entire video first. 
We provide one video frame with a 2D bounding box to specify each object instance to the annotators.
We ask annotators to draw {\it amodal} bounding boxes of each object instance and do not annotate the object instances with heavy occlusions (i.e., when less than 25\% of the object is visible).
The last type of annotation is the object motion state. 
The object is annotated as stationary only when the hands are no longer in contact with the object.
All annotations are first labeled by a group of annotators and checked by other independent annotators to ensure the quality.

\begin{figure}[t]
\includegraphics[width=1\linewidth]{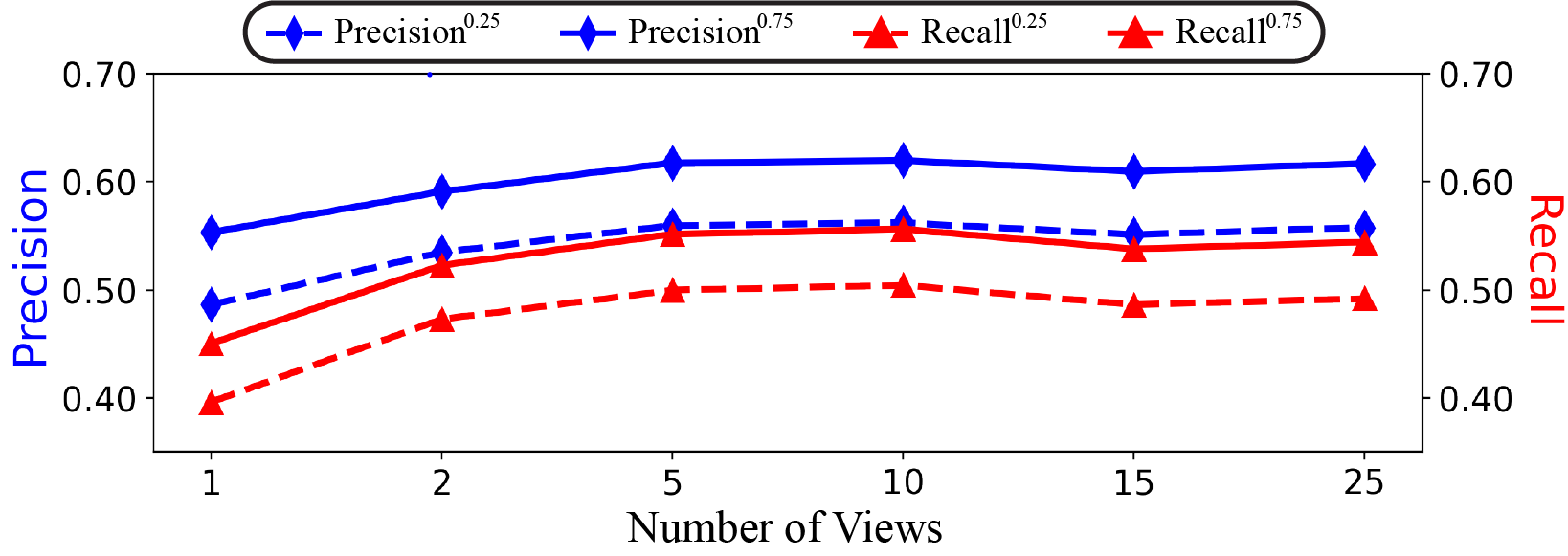}
\vspace{-6mm}
\caption{\small \textbf{Performance w.r.t number views in MVPE.} 
We run SAM+DINOv2 with different numbers of views while keeping everything else the same for a fair comparison. We find the performance saturates after using 5 views. This suggests that simply encode and average features benefit from a higher number of views (i.e., number of views from 1 to 5) but still cannot fully exploit the visual information from different views (i.e., after using 5 views).}
\vspace{-3mm}
\label{fig:num_views_ablation}
\end{figure}

\section{Additional Ablation study}
\label{sec:ablation}
This section supplements the results in the main paper with the following 4 experiments. 
We analyze the performance change w.r.t the number of views used in MVPE, memory update mechanism, feature encoder, and proposal generator of the improved baseline, i.e., SAM+DINOv2. All experiments use online enrollment (SVOE) except the number of views study.

{\bf Performance w.r.t number of views.}
Due to the architecture design of many transformer-based trackers, we only use 5 views in the benchmark experiment. In this section, we further study the relationship between the number of views and the tracking performance.
Specifically, we compare the performance of SAM+DINOv2 with 1, 2, 5, 10, 15, 25 images while keeping all other parameters the same.
As shown in Figure~\ref{fig:num_views_ablation}, the performance improves from 1 view to 5 views but quickly saturates after using 5 views. This suggests that naively encode and average features benefit from a higher number of views but still cannot fully exploit the visual information from different views.

\begin{figure}[t]
\includegraphics[width=1\linewidth]{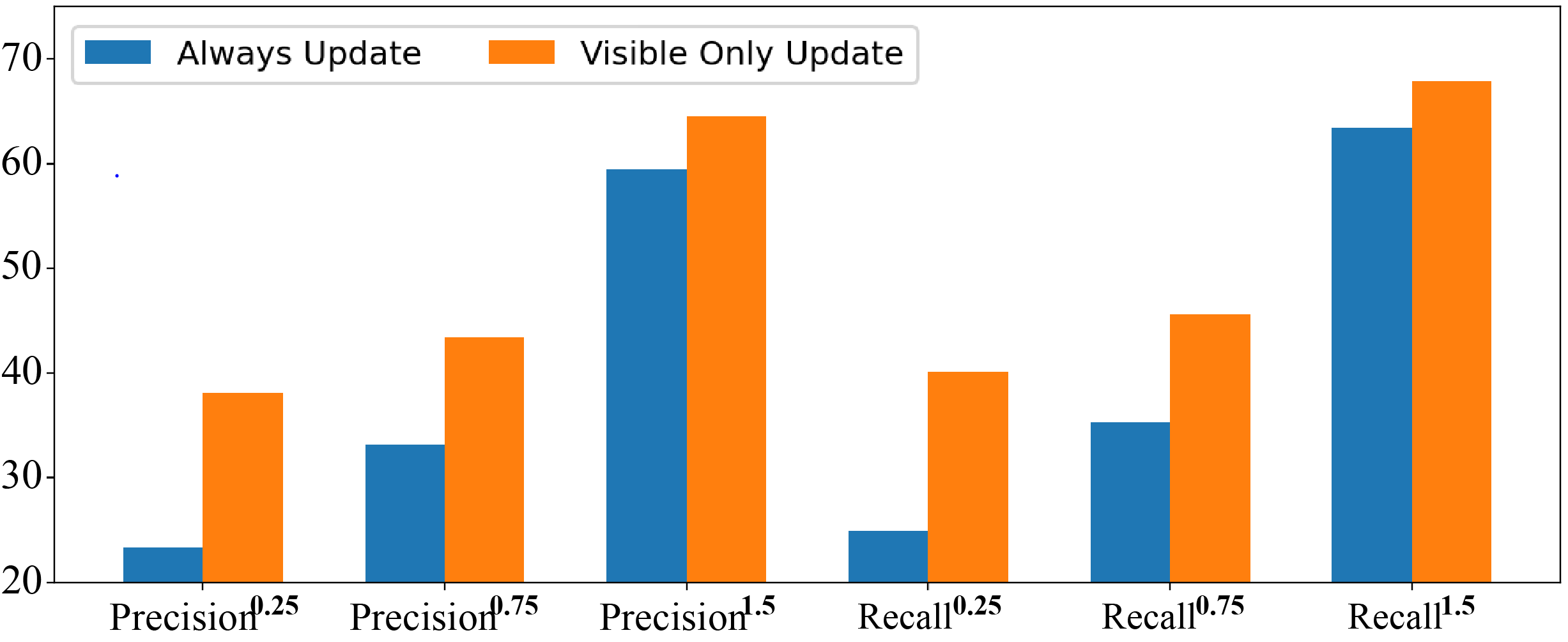}
\vspace{-6mm}
\caption{\small \textbf{Performance improvement by updating on visible only frames.} 
We control the memory update of SAM+DINOv2 by updating the memory only when the object instance is visible. We find the performance is significantly improved, indicating one of the major challenges of the baseline is to correctly update the memory with high quality predictions.
}
\vspace{-2mm}
\label{fig:visible_only_ablation}
\end{figure}

{\bf Performance improvement with visible update only.}
From the results shown in Table 2 and Table 3 in the main paper, we find identifying high quality predictions and updating the memory is the main challenge in the proposed baseline pipeline.
To further validate this idea, we control the update of memory in SAM+DINOv2 model by only updating on the visible frames. 
We extract the visible information from the 2D annotations. 
In other words, the memory for each instance is only updated on the frame where the 2D bounding box is annotated.
As shown in Figure~\ref{fig:visible_only_ablation}, updating the memory only when object instances are visible significantly improves the performance. 
Although the update timing is correct, errors from 2D predictions, depth maps and camera poses prevent the model from improving further.

\begin{figure*}[t]
\centering
\hsize=\textwidth 
\includegraphics[width=0.99\textwidth]{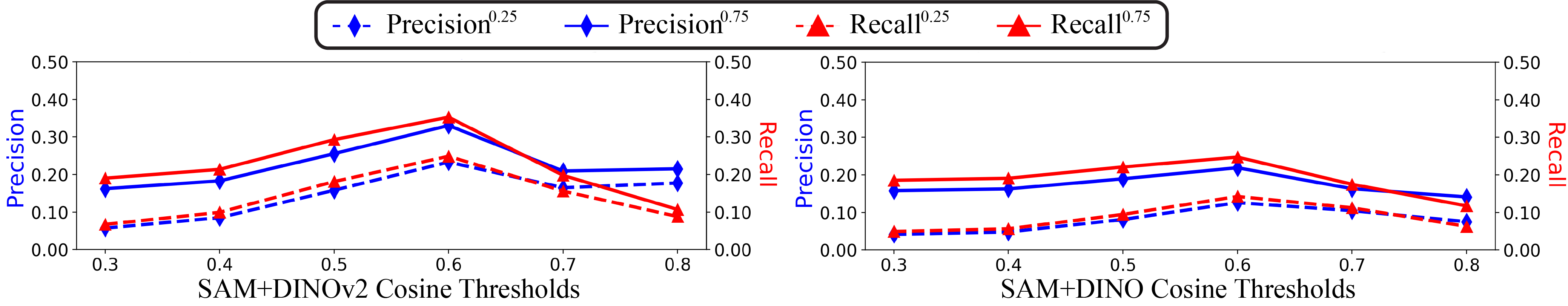}
\vspace{-2mm}
\caption{\small \textbf{Performance comparisons of different encoders at various cosine thresholds}. 
From the results, we find:
(1) {\it Stronger encoder improve the performance.} 
The best performance of SAM+DINOv2 is stronger than SAM+DINO where both models have the peak performance when the cosine threshold equals 0.6.
(2) {\it Similar performance trend w.r.t cosine similarity changes.} 
The performance of both models first improves and then gradually decreases when increasing the cosine threshold from 0.3 to 0.8.
}
\vspace{-2mm}
\label{fig:encoder_comparison}
\end{figure*}

{\bf Performance w.r.t different feature encoders.}
The top-performing baseline, i.e., SAM+DINOv2 adopts DINOv2 as the pretrained feature encoder. To further explore the performance w.r.t different large-scale feature encoders, we experiment with another state-of-the-art feature encoder, i.e., DINO~\cite{caron2021emerging}.
We plot the results using DINO and DINOv2 at different cosine thresholds in Figure~\ref{fig:encoder_comparison}.
From the results, we find:
(1) {\it Stronger encoder improves the performance.} 
The best performance of SAM+DINOv2 is stronger than SAM+DINO where both models have the peak performance when the cosine threshold equals 0.6.
(2) {\it Similar performance trend w.r.t cosine similarity changes.} 
The performance of both models first improves and then gradually decreases when increasing the cosine threshold from 0.3 to 0.8.

{
\setlength{\tabcolsep}{2.8mm}
\begin{table}[t]
\centering
\footnotesize
\caption{\small \textbf{Quantitative comparisons of different proposal generators.}
We compare the performance of SAM+DINOv2 and YOLOv7+DINOv2. To keep the comparison fair, the only differences between these models are the proposal generators. 
From the results, we find adopting YOLOv7 makes the performance slightly worse. The proposal quality from YOLOv7 is lower but runs faster.
}
\vspace{-2mm}
\scalebox{0.95}{
\begin{tabular}{cccccccc}
\toprule
\multirow{2}{*}{Proposals} & \multicolumn{3}{c}{Precision(\%)$\uparrow$} & \multicolumn{3}{c}{Recall(\%)$\uparrow$} & \multirow{2}{*}{L2$\downarrow$} \\
\cmidrule(l){2-4} \cmidrule(l){5-7}
& 0.25 & 0.75 & 1.5 & 0.25 & 0.75 & 1.5 & (m) \\
\midrule
YOLOv7 & 20.3 & 28.1 & 50.2 & 21.5 & 30.7 & 53.9 & 1.72 \\
SAM & {\bf 23.3} & {\bf 33.1} & {\bf 59.4} & {\bf 24.9} & {\bf 35.3} & {\bf 63.4} & {\bf 1.35} \\
\bottomrule
\end{tabular}}
\label{tab:proposal_comparisons}
\vspace{-4mm}
\end{table}
}

{\bf Comparisons of different proposal generators.}
Currently, the improved baseline utilizes SAM as the proposal generator. In this part, we replace SAM with 
the proposals from YOLOv7, i.e., the output before the final classification layer. The results are shown in Table~\ref{tab:proposal_comparisons}. Although the performance of YOLOv7+DINOv2 is lower compared to SAM+DINOv2, which is not surprising. The proposal quality from YOLOv7 is lower but runs faster. 
However, the current baseline approaches are not able to run in real time due to the following encoding and lifting steps. One promising direction for future work is to improve the speed of the tracking models.

\section{Datasheet}
\label{sec:datasheets for datasets}
We follow the datasheet proposed in~\cite{gebru2021datasheets} for documenting our benchmark dataset.
\vspace{1mm}
\datasheetSec{Motivation}
\vspace{1mm}

\datasheetQ{For what purpose was the dataset created?}
This dataset was created to study the problem of instance tracking in 3D from egocentric videos. 
We find current egocentric sensor data from AR/VR devices cannot support the study of our benchmark problem. 
\vspace{1mm}
\datasheetSec{Composition}
\vspace{1mm}

\datasheetQ{What do the instances that comprise the dataset represent?}
Raw egocentric video sequences, object enrollments for each object instance, and annotation files. 

\datasheetQ{How many instances are there in total?}
There are 50 video sequences with an average length of over 10K frames, 220 unique object instances with two types of enrollment information, and three types of annotations.

\datasheetQ{Does the dataset contain all possible instances or is it a sample (not necessarily random) of instances from a larger set?}
\answerYes{}
        
\datasheetQ{What data does each instance consist of?}
Please check Section 3.2 in the main paper for details.

\datasheetQ{Is there a label or target associated with each instance?}
\answerYes{} Please check Section 3.2 in the main paper for details.

\datasheetQ{Is any information missing from individual instances?}
No.

\datasheetQ{Are relationships between individual instances made explicit?}
Videos captured in the same scene share a similar surrounding environment but different activities. Object instances are related to the task performed in the video. No explicit relationships between different object instances in the same video.
        
\datasheetQ{Are there recommended data splits?}
\answerYes{} The entire benchmark dataset focuses on evaluation only. Models should be pretrained on other data sources. Please check Section 3.1 in the main paper for details.

\datasheetQ{Are there any errors, sources of noise, or redundancies in the dataset?}
\answerYes{} There are noises in camera poses and depth maps. The source of camera pose noise is from the camera localization from HoloLens2, especially under large head motion. The depth map noises are from phase wrapping. But this noise can be easily recovered with rendered depth using mesh or exploring existing unwrapping algorithms. 

\datasheetQ{Is the dataset self-contained, or does it link to or otherwise rely on external resources (e.g., websites, tweets, other datasets)?}
\answerYes{} The dataset is self-contained.

\datasheetQ{Does the dataset contain data that might be considered confidential (e.g., data that is protected by legal privilege or by doctor-patient confidentiality, data that includes the content of individuals' non-public communications)?}
\answerNo{}

\datasheetQ{Does the dataset contain data that, if viewed directly, might be offensive, insulting, threatening, or might otherwise cause anxiety?}
\answerNo{}

\datasheetQ{Does the dataset identify any subpopulations (e.g., by age, gender)?}
\answerNo{} 

\datasheetQ{Is it possible to identify individuals (i.e., one or more natural persons), either directly or indirectly (i.e., in combination with other data) from the dataset?}
\answerNo{}
We have carefully examined the data and ensure no personally identifiable information is included.

\datasheetQ{Does the dataset contain data that might be considered sensitive in any way (e.g., data that reveals racial or ethnic origins, sexual orientations, religious beliefs, political opinions or union memberships, or locations; financial or health data; biometric or genetic data; forms of government identification, such as social security numbers; criminal history)?}
\answerNo{}.

\datasheetQ{Any other comments?}
\answerNA{}
\vspace{1mm}
\datasheetSec{Collection Process}
\vspace{1mm}
\datasheetQ{How was the data associated with each instance acquired?}
The raw video sequences are collected with HoloLens2. The pre-enrollment information is captured with the iPhone 13 Pro.
The rest data, i.e, annotations and online enrollment information, are acquired from human annotators.

\datasheetQ{What mechanisms or procedures were used to collect the data (e.g., hardware apparatus or sensor, manual human curation, software program, software API)?}
The dataset is collected with open-source hl2ss~\cite{hl2ss} using HoloLens2. The pre-enrollment images are captured with the iPhone 13 Pro.
For more details please check Section 3.1 in the paper.

\datasheetQ{If the dataset is a sample from a larger set, what was the sampling strategy (e.g., deterministic, probabilistic with specific sampling probabilities)?}
\answerNA{}

\datasheetQ{Does the dataset relate to people?}
\answerYes{} The dataset includes video sequences of the first-person view of individuals performing the daily activity. 

\datasheetQ{Were any ethical review processes conducted (e.g., by an institutional review board)?}
\answerYes{} Data collection protocol was registered with the appropriate institutional review board (IRB).

\datasheetQ{Did you collect the data from the individuals in question directly, or obtain it via third parties or other sources (e.g., websites)?}
The raw video sequences are collected when the camera wearer performs the daily task.

\datasheetQ{Were the individuals in question notified about the data collection?}
\answerYes{}

\datasheetQ{Did the individuals in question consent to the collection and use of their data?}
\answerYes{}

\datasheetQ{If consent was obtained, were the consenting individuals provided with a mechanism to revoke their consent in the future or for certain uses?}
\answerNo{}

\datasheetQ{Has an analysis of the potential impact of the dataset and its use on data subjects (e.g., a data protection impact analysis) been conducted?}
\answerNo{} All annotations are on objective world states with no subjective opinions or arguments involved.

\datasheetQ{Any other comments?}
\answerNA{}
\vspace{1mm}
\datasheetSec{Preprocessing/Cleaning/Labeling}
\vspace{1mm}
\datasheetQ{Was any preprocessing/cleaning/labeling of the data done (e.g., discretization or bucketing, tokenization, part-of-speech tagging, SIFT feature extraction, removal of instances, processing of missing values)?}
\answerNo{}

\datasheetQ{Was the "raw" data saved in addition to the preprocessed/cleaned/labeled data (e.g., to support unanticipated future uses)?}
\answerYes{} We will provide both the raw data and annotations.

\datasheetQ{Is the software used to preprocess/clean/label the instances available?}
\answerNo{}

\datasheetQ{Any other comments?}
\answerNA{}
\vspace{1mm}
\datasheetSec{Uses}
\vspace{1mm}
\datasheetQ{Has the dataset been used for any tasks already?}
\answerNo{}

\datasheetQ{What (other) tasks could the dataset be used for?}
Our benchmark dataset also supports the study of other 3D scene understanding problems from egocentric videos, such as SLAM, depth estimation, and camera localization. 

\datasheetQ{Is there anything about the composition of the dataset or the way it was collected and preprocessed/cleaned/labeled that might impact future uses?}
\answerNo{} 

\datasheetQ{Are there tasks for which the dataset should not be used?}
The usage of this dataset should be limited to the scope of instance tracking in 3D and geometric scene understanding from egocentric videos.

\datasheetQ{Any other comments?}
\answerNA{}
\vspace{1mm}
\datasheetSec{Distribution}
\vspace{1mm}
\datasheetQ{Will the dataset be distributed to third parties outside of the entity (e.g., company, institution, organization) on behalf of which the dataset was created?}
\answerYes{} The dataset will be made publicly available and third parties are allowed to distribute the dataset.

\datasheetQ{How will the dataset will be distributed (e.g., tarball on website, API, GitHub)?}
The dataset will be publicly available on both Github repo and the website and stored on the cloud store, e.g., Google drive or Amazon S3.

\datasheetQ{When will the dataset be distributed?}
The full dataset will be released to the public upon acceptance of this paper. 

\datasheetQ{Will the dataset be distributed under a copyright or other intellectual property (IP) license, and/or under applicable terms of use (ToU)?}
We release our benchmark dataset and code under MIT license.

\datasheetQ{Have any third parties imposed IP-based or other restrictions on the data associated with the instances?}
\answerNo{}

\datasheetQ{Do any export controls or other regulatory restrictions apply to the dataset or to individual instances?}
\answerNo{}

\datasheetQ{Any other comments?}
\answerNA{}
\vspace{1mm}
\datasheetSec{Maintenance}
\vspace{1mm}

\datasheetQ{Is there an erratum?}
\answerNo{} 
When errors are confirmed, we will announce erratum on the platform where dataset is publicly hosted, i.e., either the Github repo or the website.

\datasheetQ{Will the dataset be updated (e.g., to correct labeling errors, add new instances, delete instances')?}
\answerYes{} 
We hope to bring more diversity to the dataset, such as more object instance and scenes. 

\datasheetQ{If the dataset relates to people, are there applicable limits on the retention of the data associated with the instances (e.g., were individuals in question told that their data would be retained for a fixed period of time and then deleted)?}
\answerNo{}

\datasheetQ{Will older versions of the dataset continue to be supported/hosted/maintained?}
\answerYes{} All versions of the dataset will be publicly available.

\datasheetQ{If others want to extend/augment/build on/contribute to the dataset, is there a mechanism for them to do so?}
Please email us if you are interested in extending or contributing to the dataset.

\datasheetQ{Any other comments?}
\answerNA{}

\section{Additional Visualizations}
\label{sec:vis}

\begin{figure*}[t] 
\centering
\hsize=\textwidth 
\includegraphics[width=0.99\textwidth]{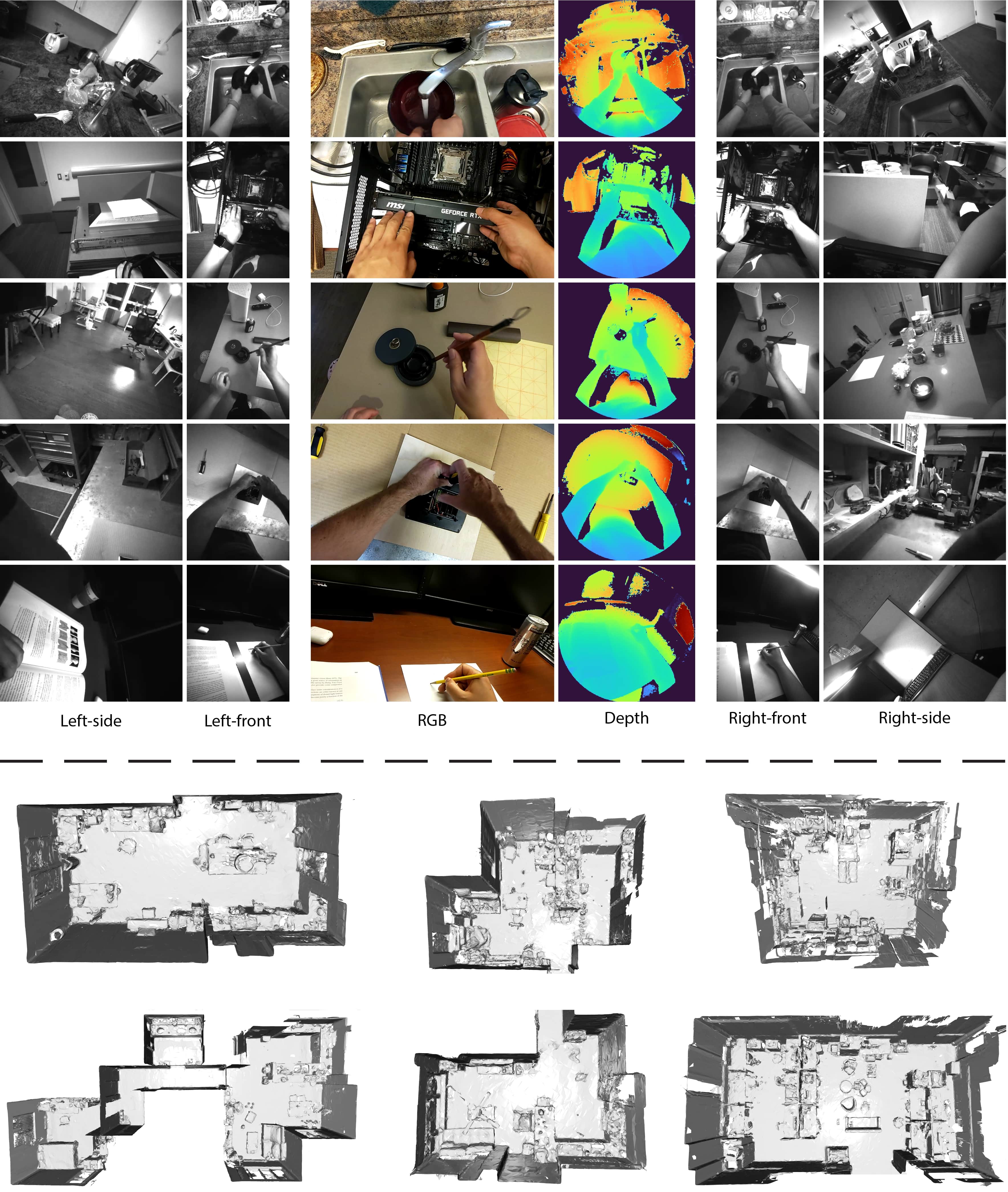}
\vspace{-2mm}
\caption{\small \textbf{2D visualizations of frames from raw video sequences (upper panel) and 3D visualizations of the capture environments (lower panel)}.
The benchmark videos record camera wearers perform naturalistic tasks in real-world scenarios, such as cooking and repairing.
Please refer to Figure~\ref{fig:data_visual} for the layout of each sensor on the HoloLens2. 
}
\label{fig:raw_data_2d_3d_visual}
\vspace{-9mm}
\end{figure*}

We include additional 2D and 3D visualizations of our benchmark dataset in Figure~\ref{fig:raw_data_2d_3d_visual}.

\end{document}

%% file: preamble.tex
\usepackage[dvipsnames]{xcolor}
\usepackage{graphics,graphicx,caption,float,subcaption,booktabs,multirow,array,color,ifthen,tabu,colortbl,dblfloatfix,url,xparse,mathtools,patchcmd,algorithm,algorithmic,amssymb,xspace,nicefrac,microtype,amsmath,amsthm,amsfonts,bm,ragged2e,tikz,stackengine,etoolbox,xpatch,enumerate,xstring,setspace,tabularx,makecell,changepage,cuted,tcolorbox}
\usepackage[utf8]{inputenc} %
\usepackage[T1]{fontenc}    %
\usepackage{pifont}%
\newcommand{\cmark}{\textcolor{ForestGreen}{\ding{51}}}%
\newcommand{\xmark}{\textcolor{BrickRed}{\ding{55}}}%

%% file: arxiv_cvpr_submission.bbl
\begin{thebibliography}{85}
\providecommand{\natexlab}[1]{#1}
\providecommand{\url}[1]{\texttt{#1}}
\expandafter\ifx\csname urlstyle\endcsname\relax
  \providecommand{\doi}[1]{doi: #1}\else
  \providecommand{\doi}{doi: \begingroup \urlstyle{rm}\Url}\fi

\bibitem[hl2(2023)]{hl2ss}
Hololens 2 sensor streaming.
\newblock \url{https://github.com/jdibenes/hl2ss}, 2023.

\bibitem[Ahmadyan et~al.(2021)Ahmadyan, Zhang, Ablavatski, Wei, and
  Grundmann]{ahmadyan2021objectron}
Adel Ahmadyan, Liangkai Zhang, Artsiom Ablavatski, Jianing Wei, and Matthias
  Grundmann.
\newblock Objectron: A large scale dataset of object-centric videos in the wild
  with pose annotations.
\newblock In \emph{Proceedings of the IEEE/CVF conference on computer vision
  and pattern recognition}, pages 7822--7831, 2021.

\bibitem[Bhat et~al.(2019)Bhat, Danelljan, Gool, and Timofte]{bhat2019learning}
Goutam Bhat, Martin Danelljan, Luc~Van Gool, and Radu Timofte.
\newblock Learning discriminative model prediction for tracking.
\newblock In \emph{Proceedings of the IEEE/CVF international conference on
  computer vision}, pages 6182--6191, 2019.

\bibitem[Bi et~al.(2020)Bi, Zhang, Mao, Deng, and Wang]{bi2020can}
Huikun Bi, Ruisi Zhang, Tianlu Mao, Zhigang Deng, and Zhaoqi Wang.
\newblock How can i see my future? fvtraj: Using first-person view for
  pedestrian trajectory prediction.
\newblock In \emph{Computer Vision--ECCV 2020: 16th European Conference,
  Glasgow, UK, August 23--28, 2020, Proceedings, Part VII 16}, pages 576--593.
  Springer, 2020.

\bibitem[Caesar et~al.(2020)Caesar, Bankiti, Lang, Vora, Liong, Xu, Krishnan,
  Pan, Baldan, and Beijbom]{caesar2020nuscenes}
Holger Caesar, Varun Bankiti, Alex~H Lang, Sourabh Vora, Venice~Erin Liong,
  Qiang Xu, Anush Krishnan, Yu Pan, Giancarlo Baldan, and Oscar Beijbom.
\newblock nuscenes: A multimodal dataset for autonomous driving.
\newblock In \emph{Proceedings of the IEEE/CVF conference on computer vision
  and pattern recognition}, pages 11621--11631, 2020.

\bibitem[Caron et~al.(2021)Caron, Touvron, Misra, J{\'e}gou, Mairal,
  Bojanowski, and Joulin]{caron2021emerging}
Mathilde Caron, Hugo Touvron, Ishan Misra, Herv{\'e} J{\'e}gou, Julien Mairal,
  Piotr Bojanowski, and Armand Joulin.
\newblock Emerging properties in self-supervised vision transformers.
\newblock In \emph{Proceedings of the IEEE/CVF international conference on
  computer vision}, pages 9650--9660, 2021.

\bibitem[Chen et~al.(2023)Chen, Liu, Zhang, Qi, and Jia]{chen2023voxelnext}
Yukang Chen, Jianhui Liu, Xiangyu Zhang, Xiaojuan Qi, and Jiaya Jia.
\newblock Voxelnext: Fully sparse voxelnet for 3d object detection and
  tracking.
\newblock In \emph{Proceedings of the IEEE/CVF Conference on Computer Vision
  and Pattern Recognition}, pages 21674--21683, 2023.

\bibitem[Cui et~al.(2022)Cui, Jiang, Wang, and Wu]{cui2022mixformer}
Yutao Cui, Cheng Jiang, Limin Wang, and Gangshan Wu.
\newblock Mixformer: End-to-end tracking with iterative mixed attention.
\newblock In \emph{Proceedings of the IEEE/CVF Conference on Computer Vision
  and Pattern Recognition}, pages 13608--13618, 2022.

\bibitem[Dai et~al.(2017)Dai, Chang, Savva, Halber, Funkhouser, and
  Nie{\ss}ner]{dai2017scannet}
Angela Dai, Angel~X Chang, Manolis Savva, Maciej Halber, Thomas Funkhouser, and
  Matthias Nie{\ss}ner.
\newblock Scannet: Richly-annotated 3d reconstructions of indoor scenes.
\newblock In \emph{Proceedings of the IEEE conference on computer vision and
  pattern recognition}, pages 5828--5839, 2017.

\bibitem[Damen et~al.(2014)Damen, Leelasawassuk, Haines, Calway, and
  Mayol-Cuevas]{damen2014you}
Dima Damen, Teesid Leelasawassuk, Osian Haines, Andrew Calway, and Walterio~W
  Mayol-Cuevas.
\newblock You-do, i-learn: Discovering task relevant objects and their modes of
  interaction from multi-user egocentric video.
\newblock In \emph{BMVC}, page~3, 2014.

\bibitem[Damen et~al.(2018)Damen, Doughty, Farinella, Fidler, Furnari, Kazakos,
  Moltisanti, Munro, Perrett, Price, et~al.]{damen2018scaling}
Dima Damen, Hazel Doughty, Giovanni~Maria Farinella, Sanja Fidler, Antonino
  Furnari, Evangelos Kazakos, Davide Moltisanti, Jonathan Munro, Toby Perrett,
  Will Price, et~al.
\newblock Scaling egocentric vision: The epic-kitchens dataset.
\newblock In \emph{Proceedings of the European Conference on Computer Vision
  (ECCV)}, pages 720--736, 2018.

\bibitem[Danelljan et~al.(2020)Danelljan, Gool, and
  Timofte]{danelljan2020probabilistic}
Martin Danelljan, Luc~Van Gool, and Radu Timofte.
\newblock Probabilistic regression for visual tracking.
\newblock In \emph{Proceedings of the IEEE/CVF conference on computer vision
  and pattern recognition}, pages 7183--7192, 2020.

\bibitem[Datta et~al.(2022)Datta, Dharur, Cartillier, Desai, Khanna, Batra, and
  Parikh]{datta2022episodic}
Samyak Datta, Sameer Dharur, Vincent Cartillier, Ruta Desai, Mukul Khanna,
  Dhruv Batra, and Devi Parikh.
\newblock Episodic memory question answering.
\newblock In \emph{Proceedings of the IEEE/CVF Conference on Computer Vision
  and Pattern Recognition}, pages 19119--19128, 2022.

\bibitem[Dimiccoli et~al.(2018)Dimiccoli, Mar{\'\i}n, and
  Thomaz]{dimiccoli2018mitigating}
Mariella Dimiccoli, Juan Mar{\'\i}n, and Edison Thomaz.
\newblock Mitigating bystander privacy concerns in egocentric activity
  recognition with deep learning and intentional image degradation.
\newblock \emph{Proceedings of the ACM on Interactive, Mobile, Wearable and
  Ubiquitous Technologies}, 1\penalty0 (4):\penalty0 1--18, 2018.

\bibitem[Droeschel et~al.(2010{\natexlab{a}})Droeschel, Holz, and
  Behnke]{droeschel2010multi}
David Droeschel, Dirk Holz, and Sven Behnke.
\newblock Multi-frequency phase unwrapping for time-of-flight cameras.
\newblock In \emph{2010 IEEE/RSJ International Conference on Intelligent Robots
  and Systems}, pages 1463--1469. IEEE, 2010{\natexlab{a}}.

\bibitem[Droeschel et~al.(2010{\natexlab{b}})Droeschel, Holz, and
  Behnke]{droeschel2010probabilistic}
David Droeschel, Dirk Holz, and Sven Behnke.
\newblock Probabilistic phase unwrapping for time-of-flight cameras.
\newblock In \emph{ISR 2010 (41st International Symposium on Robotics) and
  ROBOTIK 2010 (6th German Conference on Robotics)}, pages 1--7. VDE,
  2010{\natexlab{b}}.

\bibitem[Dunnhofer et~al.(2021)Dunnhofer, Furnari, Farinella, and
  Micheloni]{dunnhofer2021first}
Matteo Dunnhofer, Antonino Furnari, Giovanni~Maria Farinella, and Christian
  Micheloni.
\newblock Is first person vision challenging for object tracking?
\newblock In \emph{Proceedings of the IEEE/CVF International Conference on
  Computer Vision}, pages 2698--2710, 2021.

\bibitem[Dunnhofer et~al.(2023)Dunnhofer, Furnari, Farinella, and
  Micheloni]{dunnhofer2023visual}
Matteo Dunnhofer, Antonino Furnari, Giovanni~Maria Farinella, and Christian
  Micheloni.
\newblock Visual object tracking in first person vision.
\newblock \emph{International Journal of Computer Vision}, 131\penalty0
  (1):\penalty0 259--283, 2023.

\bibitem[Dwibedi et~al.(2017)Dwibedi, Misra, and Hebert]{dwibedi2017cut}
Debidatta Dwibedi, Ishan Misra, and Martial Hebert.
\newblock Cut, paste and learn: Surprisingly easy synthesis for instance
  detection.
\newblock In \emph{Proceedings of the IEEE international conference on computer
  vision}, pages 1301--1310, 2017.

\bibitem[Fathi et~al.(2012)Fathi, Hodgins, and Rehg]{fathi2012social}
Alircza Fathi, Jessica~K Hodgins, and James~M Rehg.
\newblock Social interactions: A first-person perspective.
\newblock In \emph{2012 IEEE Conference on Computer Vision and Pattern
  Recognition}, pages 1226--1233. IEEE, 2012.

\bibitem[Fernando and Herath(2021)]{fernando2021anticipating}
Basura Fernando and Samitha Herath.
\newblock Anticipating human actions by correlating past with the future with
  jaccard similarity measures.
\newblock In \emph{Proceedings of the IEEE/CVF conference on computer vision
  and pattern recognition}, pages 13224--13233, 2021.

\bibitem[Furnari and Farinella(2019)]{furnari2019would}
Antonino Furnari and Giovanni~Maria Farinella.
\newblock What would you expect? anticipating egocentric actions with
  rolling-unrolling lstms and modality attention.
\newblock In \emph{Proceedings of the IEEE/CVF International Conference on
  Computer Vision}, pages 6252--6261, 2019.

\bibitem[Garon and Lalonde(2017)]{garon2017deep}
Mathieu Garon and Jean-Fran{\c{c}}ois Lalonde.
\newblock Deep 6-dof tracking.
\newblock \emph{IEEE transactions on visualization and computer graphics},
  23\penalty0 (11):\penalty0 2410--2418, 2017.

\bibitem[Gebru et~al.(2021)Gebru, Morgenstern, Vecchione, Vaughan, Wallach,
  Iii, and Crawford]{gebru2021datasheets}
Timnit Gebru, Jamie Morgenstern, Briana Vecchione, Jennifer~Wortman Vaughan,
  Hanna Wallach, Hal~Daum{\'e} Iii, and Kate Crawford.
\newblock Datasheets for datasets.
\newblock \emph{Communications of the ACM}, 64\penalty0 (12):\penalty0 86--92,
  2021.

\bibitem[Geiger et~al.(2013)Geiger, Lenz, Stiller, and
  Urtasun]{geiger2013vision}
Andreas Geiger, Philip Lenz, Christoph Stiller, and Raquel Urtasun.
\newblock Vision meets robotics: The kitti dataset.
\newblock \emph{The International Journal of Robotics Research}, 32\penalty0
  (11):\penalty0 1231--1237, 2013.

\bibitem[Georgakis et~al.(2016)Georgakis, Reza, Mousavian, Le, and
  Ko{\v{s}}eck{\'a}]{georgakis2016multiview}
Georgios Georgakis, Md~Alimoor Reza, Arsalan Mousavian, Phi-Hung Le, and Jana
  Ko{\v{s}}eck{\'a}.
\newblock Multiview rgb-d dataset for object instance detection.
\newblock In \emph{2016 Fourth International Conference on 3D Vision (3DV)},
  pages 426--434. IEEE, 2016.

\bibitem[Grauman et~al.(2022)Grauman, Westbury, Byrne, Chavis, Furnari,
  Girdhar, Hamburger, Jiang, Liu, Liu, et~al.]{grauman2022ego4d}
Kristen Grauman, Andrew Westbury, Eugene Byrne, Zachary Chavis, Antonino
  Furnari, Rohit Girdhar, Jackson Hamburger, Hao Jiang, Miao Liu, Xingyu Liu,
  et~al.
\newblock Ego4d: Around the world in 3,000 hours of egocentric video.
\newblock In \emph{Proceedings of the IEEE/CVF Conference on Computer Vision
  and Pattern Recognition}, pages 18995--19012, 2022.

\bibitem[Hansard et~al.(2012)Hansard, Lee, Choi, and Horaud]{hansard2012time}
Miles Hansard, Seungkyu Lee, Ouk Choi, and Radu~Patrice Horaud.
\newblock \emph{Time-of-flight cameras: principles, methods and applications}.
\newblock Springer Science \& Business Media, 2012.

\bibitem[Hoda{\v{n}} et~al.(2019)Hoda{\v{n}}, Vineet, Gal, Shalev, Hanzelka,
  Connell, Urbina, Sinha, and Guenter]{hodavn2019photorealistic}
Tom{\'a}{\v{s}} Hoda{\v{n}}, Vibhav Vineet, Ran Gal, Emanuel Shalev, Jon
  Hanzelka, Treb Connell, Pedro Urbina, Sudipta~N Sinha, and Brian Guenter.
\newblock Photorealistic image synthesis for object instance detection.
\newblock In \emph{2019 IEEE international conference on image processing
  (ICIP)}, pages 66--70. IEEE, 2019.

\bibitem[Jiang and Grauman(2017)]{jiang2017seeing}
Hao Jiang and Kristen Grauman.
\newblock Seeing invisible poses: Estimating 3d body pose from egocentric
  video.
\newblock In \emph{2017 IEEE Conference on Computer Vision and Pattern
  Recognition (CVPR)}, pages 3501--3509. IEEE, 2017.

\bibitem[Kalman(1960)]{kalman1960new}
Rudolph~Emil Kalman.
\newblock A new approach to linear filtering and prediction problems.
\newblock 1960.

\bibitem[Kazakos et~al.(2019)Kazakos, Nagrani, Zisserman, and
  Damen]{kazakos2019epic}
Evangelos Kazakos, Arsha Nagrani, Andrew Zisserman, and Dima Damen.
\newblock Epic-fusion: Audio-visual temporal binding for egocentric action
  recognition.
\newblock In \emph{Proceedings of the IEEE/CVF International Conference on
  Computer Vision}, pages 5492--5501, 2019.

\bibitem[Kim et~al.(2021)Kim, O{\v{s}}ep, and Leal-Taix{\'e}]{kim2021eagermot}
Aleksandr Kim, Aljo{\v{s}}a O{\v{s}}ep, and Laura Leal-Taix{\'e}.
\newblock Eagermot: 3d multi-object tracking via sensor fusion.
\newblock In \emph{2021 IEEE International Conference on Robotics and
  Automation (ICRA)}, pages 11315--11321. IEEE, 2021.

\bibitem[Kirillov et~al.(2023)Kirillov, Mintun, Ravi, Mao, Rolland, Gustafson,
  Xiao, Whitehead, Berg, Lo, et~al.]{kirillov2023segment}
Alexander Kirillov, Eric Mintun, Nikhila Ravi, Hanzi Mao, Chloe Rolland, Laura
  Gustafson, Tete Xiao, Spencer Whitehead, Alexander~C Berg, Wan-Yen Lo, et~al.
\newblock Segment anything.
\newblock \emph{arXiv preprint arXiv:2304.02643}, 2023.

\bibitem[Kristan et~al.(2021)Kristan, Matas, Leonardis, Felsberg, Pflugfelder,
  Kamarainen, Chang, Danelljan, \v{C}ehovin Zajc, Luke\v{z}i\v{c}, Drbohlav,
  Kapyla, Hager, Yan, Yang, Zhang, Fernandez, and et. al.]{Kristan2021a}
Matej Kristan, Jir\i Matas, Ale\v{s} Leonardis, Michael Felsberg, Roman
  Pflugfelder, Joni-Kristian Kamarainen, Hyung~Jin Chang, Martin Danelljan,
  Luka \v{C}ehovin Zajc, Alan Luke\v{z}i\v{c}, Ondrej Drbohlav, Jani Kapyla,
  Gustav Hager, Song Yan, Jinyu Yang, Zhongqun Zhang, Gustavo Fernandez, and
  et. al.
\newblock The ninth visual object tracking vot2021 challenge results, 2021.

\bibitem[Kristan et~al.(2022)Kristan, Leonardis, Matas, Felsberg, Pflugfelder,
  Kamarainen, Chang, Danelljan, \v{C}ehovin Zajc, Luke\v{z}i\v{c}, Drbohlav,
  Bjorklund, Zhang, Zhang, Yan, Yang, Cai, Mayer, and Fernandez]{Kristan2022a}
Matej Kristan, Ales Leonardis, Jiri Matas, Michael Felsberg, Roman Pflugfelder,
  Joni-Kristian Kamarainen, Hyung~Jin Chang, Martin Danelljan, Luka \v{C}ehovin
  Zajc, Alan Luke\v{z}i\v{c}, Ondrej Drbohlav, Johanna Bjorklund, Yushan Zhang,
  Zhongqun Zhang, Song Yan, Wenyan Yang, Dingding Cai, Christoph Mayer, and
  Gustavo Fernandez.
\newblock The tenth visual object tracking vot2022 challenge results, 2022.

\bibitem[Lee et~al.(2012)Lee, Ghosh, and Grauman]{lee2012discovering}
Yong~Jae Lee, Joydeep Ghosh, and Kristen Grauman.
\newblock Discovering important people and objects for egocentric video
  summarization.
\newblock In \emph{2012 IEEE conference on computer vision and pattern
  recognition}, pages 1346--1353. IEEE, 2012.

\bibitem[Li et~al.(2019)Li, Cai, and Zheng]{li2019deep}
Haoxin Li, Yijun Cai, and Wei-Shi Zheng.
\newblock Deep dual relation modeling for egocentric interaction recognition.
\newblock In \emph{Proceedings of the IEEE/CVF Conference on Computer Vision
  and Pattern Recognition}, pages 7932--7941, 2019.

\bibitem[Li et~al.(2023)Li, Jain, and Shi]{li2023matting}
Jiachen Li, Jitesh Jain, and Humphrey Shi.
\newblock Matting anything.
\newblock \emph{arXiv: 2306.05399}, 2023.

\bibitem[Li et~al.(2022)Li, Cao, Liang, Liang, Chen, Zhao, and
  Feng]{li2022egocentric}
Yiming Li, Ziang Cao, Andrew Liang, Benjamin Liang, Luoyao Chen, Hang Zhao, and
  Chen Feng.
\newblock Egocentric prediction of action target in 3d.
\newblock In \emph{2022 IEEE/CVF Conference on Computer Vision and Pattern
  Recognition (CVPR)}, pages 20971--20980. IEEE, 2022.

\bibitem[Liu et~al.(2020)Liu, Tang, Li, and Rehg]{liu2020forecasting}
Miao Liu, Siyu Tang, Yin Li, and James~M Rehg.
\newblock Forecasting human-object interaction: joint prediction of motor
  attention and actions in first person video.
\newblock In \emph{Computer Vision--ECCV 2020: 16th European Conference,
  Glasgow, UK, August 23--28, 2020, Proceedings, Part I 16}, pages 704--721.
  Springer, 2020.

\bibitem[Liu et~al.(2022)Liu, Liu, Jiang, Lyu, Wan, Shen, Liang, Fu, Wang, and
  Yi]{liu2022hoi4d}
Yunze Liu, Yun Liu, Che Jiang, Kangbo Lyu, Weikang Wan, Hao Shen, Boqiang
  Liang, Zhoujie Fu, He Wang, and Li Yi.
\newblock Hoi4d: A 4d egocentric dataset for category-level human-object
  interaction.
\newblock In \emph{Proceedings of the IEEE/CVF Conference on Computer Vision
  and Pattern Recognition}, pages 21013--21022, 2022.

\bibitem[Liu et~al.(2023)Liu, Tang, Amini, Yang, Mao, Rus, and
  Han]{liu2023bevfusion}
Zhijian Liu, Haotian Tang, Alexander Amini, Xinyu Yang, Huizi Mao, Daniela~L
  Rus, and Song Han.
\newblock Bevfusion: Multi-task multi-sensor fusion with unified bird's-eye
  view representation.
\newblock In \emph{2023 IEEE International Conference on Robotics and
  Automation (ICRA)}, pages 2774--2781. IEEE, 2023.

\bibitem[Mayer et~al.(2022)Mayer, Danelljan, Bhat, Paul, Paudel, Yu, and
  Van~Gool]{mayer2022transforming}
Christoph Mayer, Martin Danelljan, Goutam Bhat, Matthieu Paul, Danda~Pani
  Paudel, Fisher Yu, and Luc Van~Gool.
\newblock Transforming model prediction for tracking.
\newblock In \emph{Proceedings of the IEEE/CVF conference on computer vision
  and pattern recognition}, pages 8731--8740, 2022.

\bibitem[Ng et~al.(2020)Ng, Xiang, Joo, and Grauman]{ng2020you2me}
Evonne Ng, Donglai Xiang, Hanbyul Joo, and Kristen Grauman.
\newblock You2me: Inferring body pose in egocentric video via first and second
  person interactions.
\newblock In \emph{Proceedings of the IEEE/CVF Conference on Computer Vision
  and Pattern Recognition}, pages 9890--9900, 2020.

\bibitem[Nguyen and Daum{\'e}~III(2019)]{nguyen2019help}
Khanh Nguyen and Hal Daum{\'e}~III.
\newblock Help, anna! visual navigation with natural multimodal assistance via
  retrospective curiosity-encouraging imitation learning.
\newblock \emph{arXiv preprint arXiv:1909.01871}, 2019.

\bibitem[Northcutt et~al.(2020)Northcutt, Zha, Lovegrove, and
  Newcombe]{northcutt2020egocom}
Curtis Northcutt, Shengxin Zha, Steven Lovegrove, and Richard Newcombe.
\newblock Egocom: A multi-person multi-modal egocentric communications dataset.
\newblock \emph{IEEE Transactions on Pattern Analysis and Machine
  Intelligence}, 2020.

\bibitem[Oquab et~al.(2023)Oquab, Darcet, Moutakanni, Vo, Szafraniec, Khalidov,
  Fernandez, Haziza, Massa, El-Nouby, et~al.]{oquab2023dinov2}
Maxime Oquab, Timoth{\'e}e Darcet, Th{\'e}o Moutakanni, Huy Vo, Marc
  Szafraniec, Vasil Khalidov, Pierre Fernandez, Daniel Haziza, Francisco Massa,
  Alaaeldin El-Nouby, et~al.
\newblock Dinov2: Learning robust visual features without supervision.
\newblock \emph{arXiv preprint arXiv:2304.07193}, 2023.

\bibitem[Pan et~al.(2023)Pan, Charron, Yang, Peters, Whelan, Kong, Parkhi,
  Newcombe, and Ren]{pan2023aria}
Xiaqing Pan, Nicholas Charron, Yongqian Yang, Scott Peters, Thomas Whelan, Chen
  Kong, Omkar Parkhi, Richard Newcombe, and Yuheng~Carl Ren.
\newblock Aria digital twin: A new benchmark dataset for egocentric 3d machine
  perception.
\newblock In \emph{Proceedings of the IEEE/CVF International Conference on
  Computer Vision}, pages 20133--20143, 2023.

\bibitem[Patel and Thakore(2013)]{patel2013moving}
Hitesh~A Patel and Darshak~G Thakore.
\newblock Moving object tracking using kalman filter.
\newblock \emph{International Journal of Computer Science and Mobile
  Computing}, 2\penalty0 (4):\penalty0 326--332, 2013.

\bibitem[Pirsiavash and Ramanan(2012)]{pirsiavash2012detecting}
Hamed Pirsiavash and Deva Ramanan.
\newblock Detecting activities of daily living in first-person camera views.
\newblock In \emph{2012 IEEE conference on computer vision and pattern
  recognition}, pages 2847--2854. IEEE, 2012.

\bibitem[Plizzari et~al.(2022)Plizzari, Planamente, Goletto, Cannici, Gusso,
  Matteucci, and Caputo]{plizzari2022e2}
Chiara Plizzari, Mirco Planamente, Gabriele Goletto, Marco Cannici, Emanuele
  Gusso, Matteo Matteucci, and Barbara Caputo.
\newblock E2 (go) motion: Motion augmented event stream for egocentric action
  recognition.
\newblock In \emph{Proceedings of the IEEE/CVF Conference on Computer Vision
  and Pattern Recognition}, pages 19935--19947, 2022.

\bibitem[Possas et~al.(2018)Possas, Caceres, and Ramos]{possas2018egocentric}
Rafael Possas, Sheila~Pinto Caceres, and Fabio Ramos.
\newblock Egocentric activity recognition on a budget.
\newblock In \emph{Proceedings of the IEEE conference on computer vision and
  pattern recognition}, pages 5967--5976, 2018.

\bibitem[Radford et~al.(2021)Radford, Kim, Hallacy, Ramesh, Goh, Agarwal,
  Sastry, Askell, Mishkin, Clark, et~al.]{radford2021learning}
Alec Radford, Jong~Wook Kim, Chris Hallacy, Aditya Ramesh, Gabriel Goh,
  Sandhini Agarwal, Girish Sastry, Amanda Askell, Pamela Mishkin, Jack Clark,
  et~al.
\newblock Learning transferable visual models from natural language
  supervision.
\newblock In \emph{International conference on machine learning}, pages
  8748--8763. PMLR, 2021.

\bibitem[Raghavan et~al.(1989)Raghavan, Bollmann, and
  Jung]{raghavan1989critical}
Vijay Raghavan, Peter Bollmann, and Gwang~S Jung.
\newblock A critical investigation of recall and precision as measures of
  retrieval system performance.
\newblock \emph{ACM Transactions on Information Systems (TOIS)}, 7\penalty0
  (3):\penalty0 205--229, 1989.

\bibitem[Rhinehart and Kitani(2017)]{rhinehart2017first}
Nicholas Rhinehart and Kris~M Kitani.
\newblock First-person activity forecasting with online inverse reinforcement
  learning.
\newblock In \emph{Proceedings of the IEEE International Conference on Computer
  Vision}, pages 3696--3705, 2017.

\bibitem[Rogez et~al.(2015)Rogez, Supancic, and Ramanan]{rogez2015first}
Gr{\'e}gory Rogez, James~S Supancic, and Deva Ramanan.
\newblock First-person pose recognition using egocentric workspaces.
\newblock In \emph{Proceedings of the IEEE conference on computer vision and
  pattern recognition}, pages 4325--4333, 2015.

\bibitem[Ryoo et~al.(2017)Ryoo, Rothrock, Fleming, and Yang]{ryoo2017privacy}
Michael Ryoo, Brandon Rothrock, Charles Fleming, and Hyun~Jong Yang.
\newblock Privacy-preserving human activity recognition from extreme low
  resolution.
\newblock In \emph{Proceedings of the AAAI conference on artificial
  intelligence}, 2017.

\bibitem[Ryoo and Matthies(2013)]{ryoo2013first}
Michael~S Ryoo and Larry Matthies.
\newblock First-person activity recognition: What are they doing to me?
\newblock In \emph{Proceedings of the IEEE conference on computer vision and
  pattern recognition}, pages 2730--2737, 2013.

\bibitem[Sch\"{o}nberger and Frahm(2016)]{schoenberger2016sfm}
Johannes~Lutz Sch\"{o}nberger and Jan-Michael Frahm.
\newblock Structure-from-motion revisited.
\newblock In \emph{Conference on Computer Vision and Pattern Recognition
  (CVPR)}, 2016.

\bibitem[Shen et~al.(2023)Shen, Zhao, Kwon, Kim, Li, and Kong]{shen2023high}
Qianqian Shen, Yunhan Zhao, Nahyun Kwon, Jeeeun Kim, Yanan Li, and Shu Kong.
\newblock A high-resolution dataset for instance detection with multi-view
  object capture.
\newblock In \emph{Thirty-seventh Conference on Neural Information Processing
  Systems Datasets and Benchmarks Track}, 2023.

\bibitem[Shi et~al.(2023)Shi, Ball, Thattai, Zhang, Hu, Gao, Shakiah, Gao,
  Padmakumar, Yang, et~al.]{shi2023alexa}
Hangjie Shi, Leslie Ball, Govind Thattai, Desheng Zhang, Lucy Hu, Qiaozi Gao,
  Suhaila Shakiah, Xiaofeng Gao, Aishwarya Padmakumar, Bofei Yang, et~al.
\newblock Alexa, play with robot: Introducing the first alexa prize simbot
  challenge on embodied ai.
\newblock \emph{arXiv preprint arXiv:2308.05221}, 2023.

\bibitem[Su and Grauman(2016)]{su2016detecting}
Yu-Chuan Su and Kristen Grauman.
\newblock Detecting engagement in egocentric video.
\newblock In \emph{Computer Vision--ECCV 2016: 14th European Conference,
  Amsterdam, The Netherlands, October 11-14, 2016, Proceedings, Part V 14},
  pages 454--471. Springer, 2016.

\bibitem[Szot et~al.(2021)Szot, Clegg, Undersander, Wijmans, Zhao, Turner,
  Maestre, Mukadam, Chaplot, Maksymets, et~al.]{szot2021habitat}
Andrew Szot, Alexander Clegg, Eric Undersander, Erik Wijmans, Yili Zhao, John
  Turner, Noah Maestre, Mustafa Mukadam, Devendra~Singh Chaplot, Oleksandr
  Maksymets, et~al.
\newblock Habitat 2.0: Training home assistants to rearrange their habitat.
\newblock \emph{Advances in Neural Information Processing Systems},
  34:\penalty0 251--266, 2021.

\bibitem[Tang et~al.(2023)Tang, Liang, Grauman, Feiszli, and
  Wang]{tang2023egotracks}
Hao Tang, Kevin Liang, Kristen Grauman, Matt Feiszli, and Weiyao Wang.
\newblock Egotracks: A long-term egocentric visual object tracking dataset.
\newblock \emph{arXiv preprint arXiv:2301.03213}, 2023.

\bibitem[Thapar et~al.(2021)Thapar, Nigam, and Arora]{thapar2021anonymizing}
Daksh Thapar, Aditya Nigam, and Chetan Arora.
\newblock Anonymizing egocentric videos.
\newblock In \emph{Proceedings of the IEEE/CVF International Conference on
  Computer Vision}, pages 2320--2329, 2021.

\bibitem[Ungureanu et~al.(2020)Ungureanu, Bogo, Galliani, Sama, Duan, Meekhof,
  St{\"u}hmer, Cashman, Tekin, Sch{\"o}nberger, et~al.]{ungureanu2020hololens}
Dorin Ungureanu, Federica Bogo, Silvano Galliani, Pooja Sama, Xin Duan, Casey
  Meekhof, Jan St{\"u}hmer, Thomas~J Cashman, Bugra Tekin, Johannes~L
  Sch{\"o}nberger, et~al.
\newblock Hololens 2 research mode as a tool for computer vision research.
\newblock \emph{arXiv preprint arXiv:2008.11239}, 2020.

\bibitem[Wang et~al.(2020)Wang, Luo, Sun, Xiong, and Zeng]{wang2020tracking}
Guangting Wang, Chong Luo, Xiaoyan Sun, Zhiwei Xiong, and Wenjun Zeng.
\newblock Tracking by instance detection: A meta-learning approach.
\newblock In \emph{Proceedings of the IEEE/CVF conference on computer vision
  and pattern recognition}, pages 6288--6297, 2020.

\bibitem[Wang et~al.(2022)Wang, Liu, Xu, Sarkar, Luvizon, and
  Theobalt]{wang2022estimating}
Jian Wang, Lingjie Liu, Weipeng Xu, Kripasindhu Sarkar, Diogo Luvizon, and
  Christian Theobalt.
\newblock Estimating egocentric 3d human pose in the wild with external weak
  supervision.
\newblock In \emph{Proceedings of the IEEE/CVF Conference on Computer Vision
  and Pattern Recognition}, pages 13157--13166, 2022.

\bibitem[Wei et~al.(2023)Wei, Bai, Zheng, Shi, and Gong]{wei2023autoregressive}
Xing Wei, Yifan Bai, Yongchao Zheng, Dahu Shi, and Yihong Gong.
\newblock Autoregressive visual tracking.
\newblock In \emph{Proceedings of the IEEE/CVF Conference on Computer Vision
  and Pattern Recognition}, pages 9697--9706, 2023.

\bibitem[Weng et~al.(2006)Weng, Kuo, and Tu]{weng2006video}
Shiuh-Ku Weng, Chung-Ming Kuo, and Shu-Kang Tu.
\newblock Video object tracking using adaptive kalman filter.
\newblock \emph{Journal of Visual Communication and Image Representation},
  17\penalty0 (6):\penalty0 1190--1208, 2006.

\bibitem[Weng et~al.(2020)Weng, Wang, Held, and Kitani]{weng20203d}
Xinshuo Weng, Jianren Wang, David Held, and Kris Kitani.
\newblock 3d multi-object tracking: A baseline and new evaluation metrics.
\newblock In \emph{2020 IEEE/RSJ International Conference on Intelligent Robots
  and Systems (IROS)}, pages 10359--10366. IEEE, 2020.

\bibitem[Wu et~al.(2023)Wu, Antonova, Kan, Lepert, Zeng, Song, Bohg,
  Rusinkiewicz, and Funkhouser]{wu2023tidybot}
Jimmy Wu, Rika Antonova, Adam Kan, Marion Lepert, Andy Zeng, Shuran Song,
  Jeannette Bohg, Szymon Rusinkiewicz, and Thomas Funkhouser.
\newblock Tidybot: Personalized robot assistance with large language models.
\newblock \emph{arXiv preprint arXiv:2305.05658}, 2023.

\bibitem[Wu et~al.(2013)Wu, Lim, and Yang]{wu2013online}
Yi Wu, Jongwoo Lim, and Ming-Hsuan Yang.
\newblock Online object tracking: A benchmark.
\newblock In \emph{Proceedings of the IEEE conference on computer vision and
  pattern recognition}, pages 2411--2418, 2013.

\bibitem[Yan et~al.(2021)Yan, Peng, Fu, Wang, and Lu]{yan2021learning}
Bin Yan, Houwen Peng, Jianlong Fu, Dong Wang, and Huchuan Lu.
\newblock Learning spatio-temporal transformer for visual tracking.
\newblock In \emph{Proceedings of the IEEE/CVF international conference on
  computer vision}, pages 10448--10457, 2021.

\bibitem[Ye et~al.(2021)Ye, Shen, Lin, Xiang, Shao, and Hoi]{ye2021deep}
Mang Ye, Jianbing Shen, Gaojie Lin, Tao Xiang, Ling Shao, and Steven~CH Hoi.
\newblock Deep learning for person re-identification: A survey and outlook.
\newblock \emph{IEEE transactions on pattern analysis and machine
  intelligence}, 44\penalty0 (6):\penalty0 2872--2893, 2021.

\bibitem[Yilmaz et~al.(2006)Yilmaz, Javed, and Shah]{yilmaz2006object}
Alper Yilmaz, Omar Javed, and Mubarak Shah.
\newblock Object tracking: A survey.
\newblock \emph{Acm computing surveys (CSUR)}, 38\penalty0 (4):\penalty0
  13--es, 2006.

\bibitem[Yin et~al.(2021)Yin, Zhou, and Krahenbuhl]{yin2021center}
Tianwei Yin, Xingyi Zhou, and Philipp Krahenbuhl.
\newblock Center-based 3d object detection and tracking.
\newblock In \emph{Proceedings of the IEEE/CVF conference on computer vision
  and pattern recognition}, pages 11784--11793, 2021.

\bibitem[Yonetani et~al.(2016)Yonetani, Kitani, and
  Sato]{yonetani2016recognizing}
Ryo Yonetani, Kris~M Kitani, and Yoichi Sato.
\newblock Recognizing micro-actions and reactions from paired egocentric
  videos.
\newblock In \emph{Proceedings of the IEEE Conference on Computer Vision and
  Pattern Recognition}, pages 2629--2638, 2016.

\bibitem[Zhao et~al.(2020)Zhao, Kong, Shin, and Fowlkes]{zhao2020domain}
Yunhan Zhao, Shu Kong, Daeyun Shin, and Charless Fowlkes.
\newblock Domain decluttering: Simplifying images to mitigate synthetic-real
  domain shift and improve depth estimation.
\newblock In \emph{Proceedings of the IEEE/CVF Conference on Computer Vision
  and Pattern Recognition}, pages 3330--3340, 2020.

\bibitem[Zhao et~al.(2021)Zhao, Kong, and Fowlkes]{zhao2021camera}
Yunhan Zhao, Shu Kong, and Charless Fowlkes.
\newblock Camera pose matters: Improving depth prediction by mitigating pose
  distribution bias.
\newblock In \emph{Proceedings of the IEEE/CVF Conference on Computer Vision
  and Pattern Recognition}, pages 15759--15768, 2021.

\bibitem[Zheng et~al.(2016)Zheng, Yang, and Hauptmann]{zheng2016person}
Liang Zheng, Yi Yang, and Alexander~G Hauptmann.
\newblock Person re-identification: Past, present and future.
\newblock \emph{arXiv preprint arXiv:1610.02984}, 2016.

\bibitem[Zhou et~al.(2022)Zhou, Luo, Luo, Liu, Pan, Cai, Zhao, and
  Lu]{zhou2022pttr}
Changqing Zhou, Zhipeng Luo, Yueru Luo, Tianrui Liu, Liang Pan, Zhongang Cai,
  Haiyu Zhao, and Shijian Lu.
\newblock Pttr: Relational 3d point cloud object tracking with transformer.
\newblock In \emph{Proceedings of the IEEE/CVF Conference on Computer Vision
  and Pattern Recognition}, pages 8531--8540, 2022.

\bibitem[Zhou et~al.(2019)Zhou, Wang, and Kr{\"a}henb{\"u}hl]{zhou2019objects}
Xingyi Zhou, Dequan Wang, and Philipp Kr{\"a}henb{\"u}hl.
\newblock Objects as points.
\newblock \emph{arXiv preprint arXiv:1904.07850}, 2019.

\bibitem[Zhu et~al.(2023)Zhu, Lai, Chen, Wang, and Lu]{zhu2023visual}
Jiawen Zhu, Simiao Lai, Xin Chen, Dong Wang, and Huchuan Lu.
\newblock Visual prompt multi-modal tracking.
\newblock In \emph{Proceedings of the IEEE/CVF Conference on Computer Vision
  and Pattern Recognition}, pages 9516--9526, 2023.

\end{thebibliography}
